%% file: EMNLP/latex/camera_ready.tex
\documentclass[11pt]{article}

\usepackage[final]{acl}
\usepackage{times}
\usepackage{latexsym}

\usepackage[T1]{fontenc}
\usepackage[utf8]{inputenc}

\usepackage{microtype}

\usepackage{inconsolata}

\usepackage{graphicx}
\usepackage{threeparttable}
\usepackage{subcaption}
\usepackage{minitoc}
\usepackage{titletoc}
\usepackage{wrapfig}
\usepackage{sidecap}
\usepackage{float} 
\usepackage{cancel}
\usepackage{booktabs}

\input{math_commands}

\newcommand{\Cat}{\text{Cat}}
\newcommand{\boldpi}{\boldsymbol{\pi}}
\newcommand{\dd}{\mathrm{d}}

\newcommand{\cfig}[2]{%
    \hyperref[#1]{\ref*{#1}#2}%
}

\definecolor{linkblue}{RGB}{80, 100, 255}

\usepackage{hyperref}
\hypersetup{
    colorlinks=true,
    linkcolor=linkblue,
    filecolor=linkblue,      
    urlcolor=linkblue,
    citecolor=linkblue,
}
\usepackage{cleveref}

\Crefformat{figure}{Fig.~(#2#1#3)}
\crefformat{figure}{Fig.~(#2#1#3)}
\Crefformat{section}{Sec.~(#2#1#3)}
\crefformat{section}{Sec.~(#2#1#3)}
\Crefformat{equation}{Eq.~(#2#1#3)}
\crefformat{equation}{Eq.~(#2#1#3)}
\Crefformat{table}{Tab.~(#2#1#3)}
\crefformat{table}{Tab.~(#2#1#3)}
\Crefformat{appendix}{Appx.~(#2#1#3)}
\crefformat{appendix}{Appx.~(#2#1#3)}
\Crefformat{algorithm}{Alg.~(#2#1#3)}
\crefformat{algorithm}{Alg.~(#2#1#3)}

\title{Language Diffusion Models are Associative Memories\\Capable of Retrieving Unseen Data}

\author{\vspace{2.5mm} \bf{
    Bao Pham$^1$ \;\;\; Mohammed J. Zaki$^1$ \;\;\; Luca Ambrogioni$^2$} \\ \vspace{2mm}
    \bf{Dmitry Krotov$^3$\;\;\; Matteo Negri$^4$
    }\\
    $^1${\selectfont \fontsize{4}{6}Rensselaer Polytechnic Institute (RPI)} \\ 
    $^2${\selectfont \fontsize{4}{6}Donders Institute for Brain, Cognition, and Behaviour, Radboud University}\\
    $^3${\selectfont \fontsize{4}{6}Dynamical Mind}\\
    $^4${\selectfont \fontsize{4}{6}Laboratoire de Physique Théorique et Modélisation, CY Cergy Paris Université}
}
\begin{document}
\maketitle
\begin{abstract}
%When do language diffusion models memorize their training data, and how to quantitatively assess their true generative regime? We address these questions by establishing that Uniform-based Discrete Diffusion Models (UDDMs) fundamentally behave as Associative Memories (AMs) \textit{with emergent creative capabilities}. The core idea of an AM is to reliably recover stored data points as \textit{memories} by establishing distinct basins around them.~Historically, models like Hopfield networks use an explicit energy function to guarantee these stable attractors.~We broaden this perspective by leveraging that energy is not strictly necessary, as basins can also be formed via conditional likelihood maximization. This conditional dynamics enables a co-existence of memory and creativity, where the UDDMs can recognize both training and unseen test sequences as fixed points and effectively denoise their corrupted versions, alongside the capability of synthesizing novel sentences. We show that, as the training dataset size increases, basins around training data points shrink while basins around unseen test data points expand, eventually becoming indistinguishable from one another. This memorization-to-generalization transition can be also detected also using the conditional entropy of generated sequences, which vanishes in the memorization regime.
Understanding when language diffusion models memorize training data versus generalize remains an open challenge. We address it by framing Uniform-based Discrete Diffusion Models (UDDMs) as Associative Memories (AMs) where both memorization and creativity coexist. Unlike Hopfield networks, which require an energy function to produce attractors, we show that attractors can be formed via conditional likelihood maximization. We observe that UDDMs treat both training and unseen test sequences as fixed points for denoising while retaining the capacity to synthesize novel texts. We demonstrate that, as training set size increases, attractor basins around training samples shrink while those around unseen test samples expand until they become indistinguishable. This transition can be detected via the conditional entropy of tokens, which vanishes during memorization. Our framework provides both a theoretical foundation and a practical metric to elucidate the generative regimes of UDDMs\footnote{Code is available at the following \href{https://github.com/Lemon-cmd/Associative-Memory-and-Language-Diffusion}{link}.}.
\end{abstract}

\section{Introduction}
\label{sec:intro}
Generative diffusion models \citep{sohl2015deep} have set new standards for image and video generation \citep{ho2020denoising, song2019generative, song2021scorebased, rombach2022high}. Yet, alongside their remarkable generative power, these models exhibit a well-documented tendency to reproduce their training data \citep{somepalli2023diffusion, somepalli2023understanding, Carlini2023, webster2023reproducible}. Although recent studies have addressed the interplay between memorization and generalization in diffusion models (DMs), they predominantly focus on the continuous image domain \citep{yoon2023diffusion, kadkhodaie2023generalization, biroli2024dynamical, kamb2024analytic, achilli2024losing, wen2024detecting, jeon2024understanding, pham2025memorization, achilli2025memorization}. Their mechanics in the discrete domain, particularly for language modeling, remain poorly investigated. 

A particularly compelling instance of these questions arises in the context of large language models. For instance, \citet{brown2020language} and \citet{kojima2022large} have found that these models are capable of few-shot and even zero-shot capabilities, referring to the model's ability to perform novel tasks with minimal or even no task-specific examples, or alternatively the possibility of retrieving appropriate responses never encountered during training. The reasons behind the emergence of these capabilities remain unclear, and the push for models capable of both factual recall and creative behaviors makes it difficult to define and assess what \textit{generalization} means in this regime.

\input{EMNLP/figures/basin_shrinkage}

\input{EMNLP/figures/recovery_rate}

Motivated by these phenomena, this work studies Uniform-based Discrete Diffusion Models (UDDMs) \citep{austin2021structured, campbell2024generative, gat2024discrete, sahoo2024simple} through the theoretical lens of Associative Memories (AMs) \cite{krotov2025modern}. Fundamentally, a generative system functions as an AM by reliably retrieving stored data points, by producing distinct basins around them \citep{hopfield1982, gardner1988space}. Historically, attempting to overload these systems with too many data points leads to a catastrophic \textit{memory blackout}, where all meaningful attractors are destroyed \cite{amit1987statistical}. However, \citet{KALAJ2025130946} has recently revealed a counterintuitive regime: overloading an AM can instead trigger a generalization phase where new attractors spontaneously form near unseen examples from the underlying data distribution, while the memorization of the training data points persists. The desired capability of language modeling of having both factual recall and creative behaviors  resembles the co-existence of memory attractors and novel attractors in AMs. For this reason, it is useful to describe the generalization phase of UDDMs from the perspective of AMs, by focusing on the retrieval of seen and unseen data points. Notably, the co-existence of these two seemingly contradicting capabilities has been studied in an analytically tractable teacher-student setting \cite{farne2026rules}.

Typically, AMs are characterized with an explicit and well-defined energy function, as seen in Hopfield networks \citep{hopfield1982, amari1972learning} and Dense Associative Memories \citep{DenseAM2016, krotovlarge}. However, this reliance on an energy function is not strictly necessary to guarantee attractor dynamics. As shown by \cite{d2025pseudo}, conditional likelihood maximization alone produces basins around the data points. Importantly, dropping the energy requirement is conceptually essential for extending the AM framework to deep feed-forward architectures. There is no reason to expect that a generic feed-forward network can be written as an energy-based model, since this interpretation restricts the class of admissible architectures due to its constraints on the symmetry of the network's weights \citep{krotov2021HAM, et, kozachkov2025neuron}. On the contrary, the conditional likelihood structure already exists in many widely used architectures (such as the Transformer \citep{vaswani2017attention}), making it a natural connection with AMs. 

\noindent \textbf{Contributions}. Leveraging conditional likelihood dynamics, our work establishes a connection between UDDMs and AMs by showing that attractor basins can be formed through conditional likelihood maximization alone (via pseudo-likelihood \cite{pseudo-likelihood}) and provide a benefit for AM. Specifically, in a simplified setting detailed in \cref{sec:am_pseudo}, we establish that maximizing conditional likelihood implicitly enforces Hebbian learning \cite{Hebb1949} on data points while maximizing their classification margins, providing a formal bridge between AM and generative models (like UDDMs) which rely on conditional likelihood. Moreover, unlike previous works which connect continuous DMs and AMs \citep{hoover2023memory, ambrogioni2023search, pham2025memorization}, we extend their results to the discrete setting of language modeling. 

This theoretical link of conditional likelihood maximization and AM allows us to interpret UDDMs as AM systems and reveals a sharp memorization-to-generalization transition governed by the training dataset size. As the training set gets bigger, basins around training examples shrink while (partial) basins around unseen test examples expand, eventually converging to a regime where novel samples become stable attractors (see Figs.~\ref{fig:basin_shrinkage} and \ref{fig:token_recovery_rate_combined}). During memorization, the UDDM fails to recognize tokens of unseen test sequences as stable points, frequently altering them during the reverse process. Once generalized, however, the model is likely to maintain these tokens, preserving them if unperturbed and successfully recovering them from partially corrupted sequences via the reverse process (see Figs.~\cfig{fig:basin_shrinkage}{A} and \cfig{fig:unrecovered_recovered}{A} for visual examples of test and training points' token recovery). 

Empirically, we detect this transition using \textit{token recovery rates} and \textit{conditional entropy} as complementary probes for token stability and the memorization-to-generalization transition. Crucially, token-level conditional entropy distinguishes between different token recovery behaviors, with successfully recovered tokens exhibiting near-zero entropy (see Fig.~\cfig{fig:unrecovered_recovered}{}). Meanwhile, sequence-level conditional entropy serves as a practical metric for detecting generalization, as the entropy distributions of training and synthetic sequences align perfectly once the transition is reached (see Fig.~\ref{fig:seq_cond}). Notably, conditional entropy has also been found to be crucial for scaling laws in autoregressive language models \cite{cagnetta2026deriving}, suggesting the versatility and usefulness of this metric in language modeling.

Finally, we validate our proposed metrics via the typical measurement of perplexity using an external critic model as commonly done in language modeling \cite{radford2019language}. We observe a strong alignment between these metrics, not just in the detection of memorization-to-generalization transition point, but also in the estimation of sample quality (see Fig.~\ref{fig:perplexity}). The success of this validation indicates that our metrics are effective for assessing the generative capabilities of UDDMs, while also providing a principled geometric interpretation of such capabilities.

\section{Uniform-State Discrete diffusion} \label{sec:prelim}
Consider a clean token $\rvx \in \mathcal{V}$ drawn from the data distribution $q_\text{data}$ with the vocabulary $\mathcal{V} = \{ \rvx \in \{0, 1\}^K : \sum^K_{i =1} \rvx_i =1 \}$. In the DDM framework, $q_\text{data}$ is mapped into a simple distribution through a sequence of Markov states via a forward process that is somewhat akin to the continuous diffusion framework \citep{austin2021structured, gat2024discrete, sahoo2024simple, sahoo2025duality}:
\begin{equation}
    \rvz_t \sim q_t (\rvz_t | \rvx; \alpha_t) = \Cat(\rvz_t; \alpha_t \rvx + (1 - \alpha_t) \boldpi),
    \label{eqn:discrete-forward-process}
\end{equation}
where $\boldpi \in \Delta$, $\Cat(\cdot)$ denotes categorical distribution, and $\Delta$ denotes $K$-simplex. Here, $\rvz_t$ denotes the perturbed token at a time $t \in (0, 1]$, where $\rvz_0 = \rvx$. The diffusion parameter $\alpha_t \in [0, 1]$ is a strictly decreasing $t$-dependent function with the boundary conditions: $\alpha_{t=0} \approx 1$ and $\alpha_{t=1} \approx 0$.

\input{EMNLP/figures/unrecovered_recovered}

In UDDM, as shown by \citet{austin2021structured} and \citet{campbell2024generative}, the true reverse posterior of a previous timestep $s < t$ corresponding to the forward process (\ref{eqn:discrete-forward-process}) is
\begin{equation}
    \rvz_s \sim q_{s | t}(\rvz_s | \rvz_t, \rvx) = \text{Cat} \Big (\rvz_s; g(\rvz_t, \rvx) \Big ), %= \text{Cat} \Bigg (\rvz_s ; \frac{K \alpha_t \rvz_t \odot \rvx + (\alpha_{t|s} -\alpha_t)\rvz_t}{K \alpha_t \langle \rvz_t, \rvx \rangle + (1 - \alpha_t)} + \frac{(\alpha_s - \alpha_t)\rvx + (1 - \alpha_{t | s}) \frac{\mathbf{1}}{K}}{K \alpha_t \langle \rvz_t, \rvx \rangle + (1 - \alpha_t)} \Bigg ), 
    \label{eqn:discrete-reverse-process}
\end{equation}
%where $\langle \cdot, \cdot \rangle$ denotes the dot product, $\odot$ denotes Hadamard product, the relative diffusion parameter is $\alpha_{t|s} =\frac{\alpha_t}{\alpha_s}$, and we have a uniform prior over $\mathcal{V} (\boldpi = \mathbf{1} / {K})$ \citep{sahoo2024simple, sahoo2025duality}.
where $g(\rvz_t, \rvx)$ parameterizes the categorical transition probabilities for the reverse step by combining $\rvz_t$, $\rvx$, and uniform noise according to the schedule coefficients. Here, we have a uniform prior over $\mathcal{V} (\boldpi = \mathbf{1} / {K})$ \citep{sahoo2024simple, sahoo2025duality}.

Since the true clean token $\rvx$ is unknown during the generative process (\ref{eqn:discrete-reverse-process}), the approximate reverse posterior is defined as $p^\theta_{s|t}(\rvz_s | \rvz_t) = q_{s | t} (\rvz_s | \rvz_t, \rvx = \rvx_\theta(\rvz_t, t))$. The neural network $\rvx_\theta(\rvz_t, t) \approx \rvx$ is trained to predict the clean token $\rvx$ at any time $t$ according to \cref{eqn:discrete-reverse-process}, which we can formally define as the conditional probability distribution $p_\theta (\rvx | \rvz_t) = \rvx_\theta (\rvz_t, t)$. The network's parameters $\theta$ are optimized via the Negative Evidence Lower Bound (NELBO) objective \citep{austin2021structured, sahoo2024simple}: 
\begin{equation}
\mathcal{L}_\text{\tiny NELBO}=
    \mathbb{E}_q \Big [
%    \underset{\mathcal{L}_\text{reconstruction}}{\underbrace{- \log p_\theta (\rvx | \rvz_0)}} + \underset{\mathcal{L}_\text{diffusion}}{\underbrace{\sum_{s < t} D_\text{KL} \big [ q(\rvz_s | \rvz_t , \rvx) \, || \, p_\theta (\rvz_s | \rvz_t ) \big ]}}
%    \Big ] + \underset{\mathcal{L}_\text{prior}}{\underbrace{D_\text{KL} \big [ q(\rvz_1 | \rvx) \, || \,  p_\theta (\rvz_1) \big ]}}.
\mathcal{L}_\text{\tiny reconstruction} + \mathcal{L}_\text{\tiny diffusion} + \mathcal{L}_\text{\tiny prior} \Big ]
    \label{eqn:nelbo}
\end{equation}

By framing the network's prediction as a categorical probability distribution, the cross-entropy terms within $\mathcal{L}_\text{NELBO}$ naturally enforce a finite classification margin and therefore store training data points. As we will explore next, this reliance on cross-entropy provides a mathematical link between discrete diffusion dynamics and AM systems. Lastly, the form of UDDMs we used is based on \cite{sahoo2025duality}. Please see Appx.~(\ref{sec:duo}) for more details on this form of UDDMs, \cref{eqn:discrete-reverse-process}, and NELBO (\ref{eqn:nelbo}).

\input{EMNLP/latex/section3}

\input{EMNLP/figures/seq_cond}

\section{Methods}
\label{sec:results}
As shown in \citet{d2025pseudo} and \citet{KALAJ2025130946},
the AM produced by conditional-likelihood (\ref{eqn:pseudo-likelihood}) is capable of generalization: when trained on a sufficiently large dataset, the system creates new attractors that are strongly correlated with both training and test examples. In this section, taking inspirations from these results, we can ask whether similar regimes of memorization and generalization, found in AMs, can appear in UDDMs. Consequently, we design three experiments of increasing realism from the UDDM perspective, progressively bridging the gap between the AM retrieval setting and the standard generative process of a UDDM. Across these experiments, we explore the memorization-to-generalization transition as a function of the training dataset size and model scale.

%To understand how far the AM analogy goes, we design three experiments of increasing realism from the UDDM perspective, progressively bridging the gap between the AM retrieval setting and the standard generative process of a UDDM. Across these experiments, we explore the memorization-to-generalization transition as a function of the training dataset size and model scale.

%\textbf{Experiment 1: Deterministic retrieval from reference examples}. In Fig.~(\ref{fig:basin_shrinkage}) we report the setting closest to a classical AM: we initialize the reverse process from corrupted training and test examples, and replace the standard stochastic sampling with a deterministic greedy dynamics. At each step, we take the $\argmax$ of the conditional probabilities rather than sampling from them. In this setting, the perturbation level is decoupled from the diffusion time, which is always set to $t=1$ while the fraction of corrupted input tokens is varied. Here, we measure the token recovery rate as a function of the input noise level, while emulating the zero-temperature retrieval dynamics of an AM. 

\smallskip \noindent
\textbf{Experiment 1: Deterministic retrieval from reference examples}. To emulate the zero-temperature retrieval dynamics of an AM without the confounding effects of stochastic generation noise, we establish a specialized baseline setting. As reported in Fig.~(\ref{fig:basin_shrinkage}), we initialize the reverse process from corrupted training and test examples, and replace the standard stochastic sampling with deterministic greedy dynamics. At each step, we take the $\argmax$ of the conditional probabilities rather than sampling from them. In this setting, the perturbation level is decoupled from the diffusion time, which is always set to $t=1$ while the fraction of corrupted input tokens is varied, allowing us to cleanly measure the token recovery rate as a function of the input noise level. 

\medskip \noindent 
\textbf{Experiment 2: Stochastic retrieval from reference examples}. To understand how the standard diffusion dynamics alter the stability and structural boundaries of these basins under realistic operating conditions, we introduce stochasticity back into the UDDMs. In Figs.~(\ref{fig:token_recovery_rate_combined}) and (\ref{fig:unrecovered_recovered}) we retain the initialization from training and test examples, but restore the standard stochastic reverse dynamics. The perturbation level is now tied to the diffusion time $t$ as in the typical UDDM scheme, where each value of $t$ determines both the fraction of corrupted tokens and the time step at which the reverse process is initialized, providing a dynamic look at token recovery across varying levels of input noise.

%In Figs.~(\ref{fig:token_recovery_rate_combined}) and (\ref{fig:unrecovered_recovered}) we retain the initialization from training and test examples, but restore the standard stochastic reverse dynamics. The perturbation level is now tied to the diffusion time $t$ as in the typical UDDM scheme. Each value of $t$ determines both the fraction of corrupted tokens and the time step at which the reverse process is initialized. We again measure the token recovery rate across varying levels of input noise.

\medskip \noindent 
\textbf{Experiment 3: Standard generative process}. Finally, we run the full standard reverse dynamics starting from \textit{noise}, independent of any reference sequences, e.g., training or test sequences, in Figs.~(\ref{fig:seq_cond}) and (\ref{fig:perplexity}). Because ground-truth samples are unavailable, token recovery rate is no longer a viable metric. Instead, we use conditional entropy (\ref{eqn:cond-ent}) as our metric. In Fig.~(\ref{fig:perplexity}) we validate the performance of UDDM across the memorization-to-generalization transition by measuring the perplexity of the generated sequences with GPT-2 Large \cite{radford2019language} used as the critic model.

%In Figs.~(\ref{fig:seq_cond}) and (\ref{fig:perplexity}) we run the full standard reverse dynamics starting from a random initial condition, independent of any reference sequences (e.g., training or test examples). 
% Because ground-truth configurations are unavailable, token recovery rate is no longer a defined metric. Instead, we use conditional entropy as a metric, as defined below. 

%\textbf{Validation}. 
%In Fig.~(\ref{fig:perplexity}) we validate the performance of UDDM's across the memorization-to-generalization transition by measuring the perplexity of the generated sequences with GPT-2 Large \cite{radford2019language} being used as the reference or critic model.  Please see Appx.~(\ref{sec:experimental-details}) for further details of these experiments. 

We analyze these experiments on the LM1B dataset \citep{chelba2013one} via two metrics, as a function of the training dataset size: \textit{token recovery rate} and \textit{conditional entropy}, which serve as proxies for the stability and geometry of the attractors, respectively. For more details and results, please refer to Appx.~(\ref{sec:experimental-details}) and Appx.~(\ref{sec:add-results}), respectively.

\medskip \noindent 
\textbf{Corrupt Token Recovery}. We define the corrupt token recovery rate as the accuracy with which the model recovers a target or original sequence of length $L$, either a training or a test example, via applying the reverse process on a noisy sequence defined at time $t \in (0, 1]$: 
\begin{equation}
    R(\rvx^{1 : L}, \hat{\rvx}^{1:L}) = \frac{1}{|\mathcal{M}|} \sum_{j \in \mathcal{M}} \delta (\rvx^j, \hat{\rvx}^j \big ),
\label{eqn:token-recovery-rate}
\end{equation}
where $\rvx^{1 : L}$ is the original sequence, $\hat{\rvx}^{1 : L}$ is the recovered sequence after running a denoising process (either greedy or stochastic), $\mathcal{M} = \{ j : \delta(\rvx^j, \rvz_t^j) = 0\}$ is the set of indices in the input sequence $\rvz_t^{1: L}$ denoting the positions where the tokens of $\rvx^{1:L}$ have been changed after applying the forward process (\ref{eqn:discrete-forward-process}), and $\delta(\cdot, \cdot)$ denotes the Kronecker delta function. 

%While token recovery rates indicate whether the system successfully returns to an attractor, UDDMs also provide direct access to conditional likelihood, allowing us to probe some information about the local geometry and sharpness of these basins. In the more common setting where AM have an energy landscape, the sharpness of the basin dictates retrieval dynamics \citep{DenseAM2016, KrotovFrontiers, krotov2025modern}. Low entropy implies more deterministic attractors (\textit{memorization}), while high entropy signals a flatter landscape with distributed probability mass \citep{biroli2024dynamical}. This flatness facilitates \textit{generalization}, where basins widen and merge to capture the underlying data manifold, enabling the synthesis of novel patterns \citep{pham2025memorization}. See Appx.~(\ref{sec:entropy_curvature}) for further discussion on the relationship between conditional entropy and energy curvature. 

\medskip \noindent \textbf{Conditional Entropy}. 
While token recovery rates indicate whether the system successfully returns to an attractor, UDDMs also provide direct access to conditional likelihood, allowing us to probe the local geometry and sharpness of these basins. Here, the conditional entropy of an individual token $\rvx^\ell$ in a sequence, given its associated perturbed sequence $\rvz_t^{1 : L}$ at some time $t$ is defined as:
\begin{equation}
\begin{split}
&\mathcal{H}(\rvx^\ell | \rvz_t^{1 : L}) \\%= - \sum^K_{k = 1} \big[p_\theta (\rvx^\ell \; |\; \rvz_t^{1 : L}) \big]_k \cdot \, \log \big[p_\theta (\rvx^\ell \; | \; \rvz_t^{1 : L}) \big]_k \\ 
&=-\sum^K_{k = 1} \big [\rvx_\theta^\ell(\rvz_t^{1 : L}, t) \big ]_k \log \big [ \rvx_\theta^\ell(\rvz_t^{1 : L}, t) \big ]_k,
\end{split}
\label{eqn:cond-ent}
\end{equation}
where $\rvx_\theta^\ell(\rvz_t^{1: L}, t)$ is the output of the diffusion transformer for position $\ell$ of the sequence over a vocabulary of size $K$. Meanwhile, the conditional entropy of a sequence is the sum of each token's conditional entropy for all positions from 1 to $L$, i.e., $\mathcal{H}(\rvx^{1: L} \; | \; \rvz_t^{1: L}) = \sum^L_{\ell = 1} \mathcal{H}(\rvx^\ell \; | \; \rvz_t^{1 : L})$. Please refer to Appx.~(\ref{sec:entropy_curvature}) for our discussion on the relationship between conditional entropy and energy curvature (or sharpness).

\input{EMNLP/figures/perplexity}

%(\ref{fig:entropy_over_time}),

% This experiment evaluates the UDDM's standard generative behavior, demonstrating that conditional entropy serves as a practical probe for the memorization-to-generalization transition, eliminating the need to explicitly verify whether generated samples are duplicates of the training dataset as its size increases. To validate the performance of UDDM's across the memorization-to-generalization transition measure the perplexity of the generated sequences, shown in \cref{fig:perplexity}, where GPT-2 Large \cite{radford2019language} is used as the reference or critic model. Please see Appx.~(\ref{sec:experimental-details}) for further details of these three experiments.

\section{Results}
Our empirical analysis demonstrates that, as the training set size increases, basins around training examples shrink and basins around test examples expand (see Fig.~\ref{fig:basin_shrinkage}). Changes in the structure of basins are reflected in token recovery rates: for larger datasets the token recovery rate for training examples drops (see Fig.~\ref{fig:token_recovery_rate_combined}, \textit{solid} lines), while the recovery rate for unseen test samples increases (\textit{dashed} lines in Fig.~\ref{fig:token_recovery_rate_combined}). For a certain size onward, the retrieval behaviors of both seen and unseen sequences converge to an identical recovery rate, mirroring the convergence of training and test basins and completing the transition from memorization to generalization. Scaling up the model size delays this transition, consistently with continuous DMs \cite{yoon2023diffusion, pham2025memorization}.  

For these results, please note that a high recovery rate on test sequences does not imply that a basin has formed around them, since it could also be explained just by a correct conditional prediction. We keep the naming of `basin' for two reasons. First, the denoising is still non-trivial: the rate remains well above the injected noise, so the dynamics returns close to the original sequence rather than to an unrelated one. Second, a sentence admits many valid completions, and a rate below $100\%$ is what we should expect, since valid alternatives are counted as unrecovered tokens. Both are compatible with the attractor being flat, with the flat directions corresponding to the equivalent formulations of a given sentence, which also accounts for the dependence on the noise level: without corruption the greedy reverse process initialized on a test sequence stays on it, giving $100\%$ recovery, while any finite corruption drops it below $100\%$ (see Fig.~\ref{fig:basin_shrinkage}B), because the state can be moved to a different point of the same basin. Consistently, the unrecovered tokens in the generalization phase are usually grammatically and semantically valid (see Figs.~\ref{fig:basin_shrinkage}A, \ref{fig:train_recovery_examples}, and \ref{fig:test_recovery_examples}).

% Our empirical analysis demonstrates that as the training dataset size increases, the token recovery rate for training examples drops (\cref{fig:token_recovery_rate_combined}, \textit{solid} lines), reflecting a shrinkage of the basins around such points shown in Fig.~(\cfig{fig:basin_shrinkage}{A}). Meanwhile, the recovery rate for unseen test samples steadily improves (\textit{dashed} lines in Fig.~\ref{fig:token_recovery_rate_combined}), signaling an expansion of their respective basins. For larger training sets, the retrieval behaviors of both seen and unseen sequences converge to an identical recovery rate, completing the transition from memorization to generalization where novel samples become stable attractors. Scaling up the model parameter size delays this phase transition, prolonging the memorization regime by requiring a higher data volume to trigger convergence, a property consistent with continuous DMs \cite{yoon2023diffusion, pham2025memorization}.  

Moreover, as shown in \cref{fig:unrecovered_recovered}, successful token recovery is characterized by near-zero conditional entropy, whereas unrecovered tokens exhibit higher and non-zero conditional entropy value. Interestingly, a distinct fraction of highly stable and low-entropy tokens persists even at the later generalization phase, suggesting that there are tokens which are very stable and less likely to change through the generative process. Using the average \textit{sequence-level} conditional entropy of training and generated sequences, we observe a conditional \textit{entropy gap} at the intermediate regime of the transition (see Fig.~\ref{fig:seq_cond}): at very low fractions of the training set, the conditional entropy distributions of training and generated sequences coincide and peak near zero. In the intermediate regime, these distributions however separate from each other until the training set size is large enough, and once again overlap each other (but at \textit{non-zero} values of conditional entropy). 
%Perplexity benchmarks using GPT-2 Large validate these phase transitions (see Fig.~\ref{fig:perplexity}). In the memorization regime, perplexity remains artificially low due to the duplication of training data points. As the size of training set scales, perplexity peaks at the boundary and subsequently converges to a lower value (but higher than the initial low value) upon entering the generalization phase. Simultaneously overlapping with the trends in token recoverability, scaling the model size yields superior final perplexity values. The tight structural alignment, demonstrated by the  between our geometric probes and external perplexity benchmarks across the entire dataset scale demonstrates that token recovery rate serves as a robust, self-contained metric to diagnose the generative regimes of discrete diffusion models without requiring an external critic.  

Perplexity benchmarks using GPT-2 Large validate these phase transitions (see Fig.~\ref{fig:perplexity}): we observe that the region where perplexity increases (to the \textit{left} of the transition) corresponds to the memorization phase, which suggest that the perplexity is initially low due to the model's duplication of training sequences. Conversely, the region where perplexity decreases (to the \textit{right} of the transition) corresponds to the generalization phase where basins around test sequences start forming, suggesting that these novel basins are related to the generative capabilities of the model. 

The correspondence between token recovery rate in \cref{fig:token_recovery_rate_combined} and perplexity in \cref{fig:perplexity} also holds for the rightmost region where the perplexity plateaus: the length of the plateaus is similar in both figure. The plateaus of lower perplexity correspond to the converging token recovery rates, see \cref{fig:ppl_combined} in Appx.~(\ref{sec:add-results}) for a combined version of Figs.~(\ref{fig:token_recovery_rate_combined}) and (\ref{fig:perplexity}). This alignment suggests that token recovery rate captures the same underlying phenomenon: it identifies the generative regime without an external model and offers a geometrical interpretation, where generalization is defined by the evolution of basins around training and test sequences.

Note that recovery under denoising is not the same as training-data extraction \citep{Carlini2023, webster2023reproducible}. Our metrics probe the stability of the dynamics near a given sequence, and carry no information on the probability of reaching a specific training or test sequence starting from noise. The non-monotonic perplexity of Fig. (\ref{fig:perplexity}) is only indirect evidence in this direction: the low perplexity at small fractions is compatible with the model duplicating training sequences, while the plateau after the transition we can only interpret as good in-distribution generation.

\section{Conclusion}
\label{sec:conclusion}
%By interpreting UDDMs through the lens of AMs, we show that an explicit energy function is not necessary to guarantee attractor dynamics:~basins can form through conditional likelihood maximization alone, which naturally extends the AM framework to architectures like the diffusion transformer. This perspective reveals a memorization-to-generalization transition governed by training dataset size. As the dataset grows, basins around training examples shrink while basins around unseen test examples expand, until becoming indistinguishable from each other. This aspect allows for the recovery of training and unseen test sequences from their perturbations, emphasizing the co-existence of memory and creativity in UDDMs. 

%Moreover, token recovery rate and conditional entropy serve as complementary geometrical probes of this transition, as they describe basins by measuring the stability of token sequences with respect to their perturbations. In the memorization regime conditional entropy vanishes, while in the generalization regime it remains finite, producing a measurable entropy gap between training and synthetic sequences, which narrows as the model's size scales. Larger models also delay the transition, requiring more data before generalization emerges, and both metrics stabilize past that point, possibly suggesting diminishing returns from further training. 

By framing UDDMs as AMs, we show that an explicit energy function is not required to guarantee attractor dynamics. Instead, basins can form through conditional likelihood maximization alone, which reveals a sharp memorization-to-generalization transition governed by the training set size. As training dataset scales up, basins around training examples shrink while basins around unseen test sequences expand until they become indistinguishable. This evolution allows the model to recover both seen and unseen sequences from perturbations, showing a form of co-existence of memory and creativity. 

Furthermore, token recovery rate and conditional entropy serve as complementary geometric probes of this transition: conditional entropy vanishes during memorization but remains finite during generalization, providing a measurable entropy gap between training and synthetic sequences that narrows with model scale. Larger models delay this transition, requiring more data to generalize before both metrics stabilize past the phase boundary. In this picture the attractors in the generalization phase are flat, where the flat directions correspond to equivalent formulations of a sentence. We leave the study of these directions, which requires assessing semantic equivalence, to future work.

% Finally, as noted in our results, a distinct fraction of highly stable, low-entropy tokens persists even during the generalization phase where novel text is being synthesized. We hypothesize that these specific tokens correspond to rigid syntactic structures, highly predictable function words, or common local idioms that must remain stable for the model to preserve linguistic fluency. Because natural language relies heavily on these “stable words,” certain tokens naturally act as highly stable micro-attractors that are extremely unlikely to change through the generative process, even when the broader sequence represents an entirely novel or different semantic content from the original sequence. Nonetheless, this leaves an interesting question for us and the community to explore.

Finally, a distinct fraction of highly stable, low-entropy tokens persists even during the generalization phase, when novel text is being synthesized. We hypothesize that these tokens correspond to rigid syntactic structures, predictable function words, or common idioms that must remain stable to preserve linguistic fluency, and that act therefore as micro-attractors within an otherwise novel sequence. This leaves an interesting question for us and the community to explore.

\section*{Limitations}
This work opens some directions for future investigation. First, how good the new proposed metrics are in practically relevant cases remains an open question: while we showed a promising consistent picture with the perplexity measurement, an extensive analysis of how token recovery rate and conditional entropy correlate with standard evaluation metrics would be needed to consolidate their validity as proxies for generalization. Second, UDDMs provide a convenient and tractable setting for this study, but extending these ideas to regimes where factual recall appears in practice would require significantly larger models, up to and including large language models, which would introduce conceptual and practical challenges beyond the scope of this work.

\section*{Acknowledgment}
Matteo Negri thanks Lenka Zdeborová for insightful discussions on the role of factual recall in generative models. Bao Pham and Mohammed Zaki acknowledge funding provided by the RPI-IBM Future of Computing Research Collaboration. Part of the work of Dmitry Krotov was performed during his employment at IBM.

\bibliography{sn-bibliography}

\clearpage
\appendix
\input{EMNLP/latex/appendix_revised}

\end{document}

%% file: math_commands.tex
\usepackage{amsmath,amsfonts,bm}

\def\eqref#1{equation~\ref{#1}}
\def\1{\bm{1}}

\def\rvs{{\mathbf{s}}}

\def\rvw{{\mathbf{w}}}
\def\rvx{{\mathbf{x}}}
\def\rvy{{\mathbf{y}}}
\def\rvz{{\mathbf{z}}}

\DeclareMathAlphabet{\mathsfit}{\encodingdefault}{\sfdefault}{m}{sl}
\SetMathAlphabet{\mathsfit}{bold}{\encodingdefault}{\sfdefault}{bx}{n}

\DeclareMathOperator*{\argmax}{arg\,max}

%% file: EMNLP/figures/basin_shrinkage.tex
\begin{figure*}[!ht]
     \centering
     \includegraphics[width=1.0\textwidth]{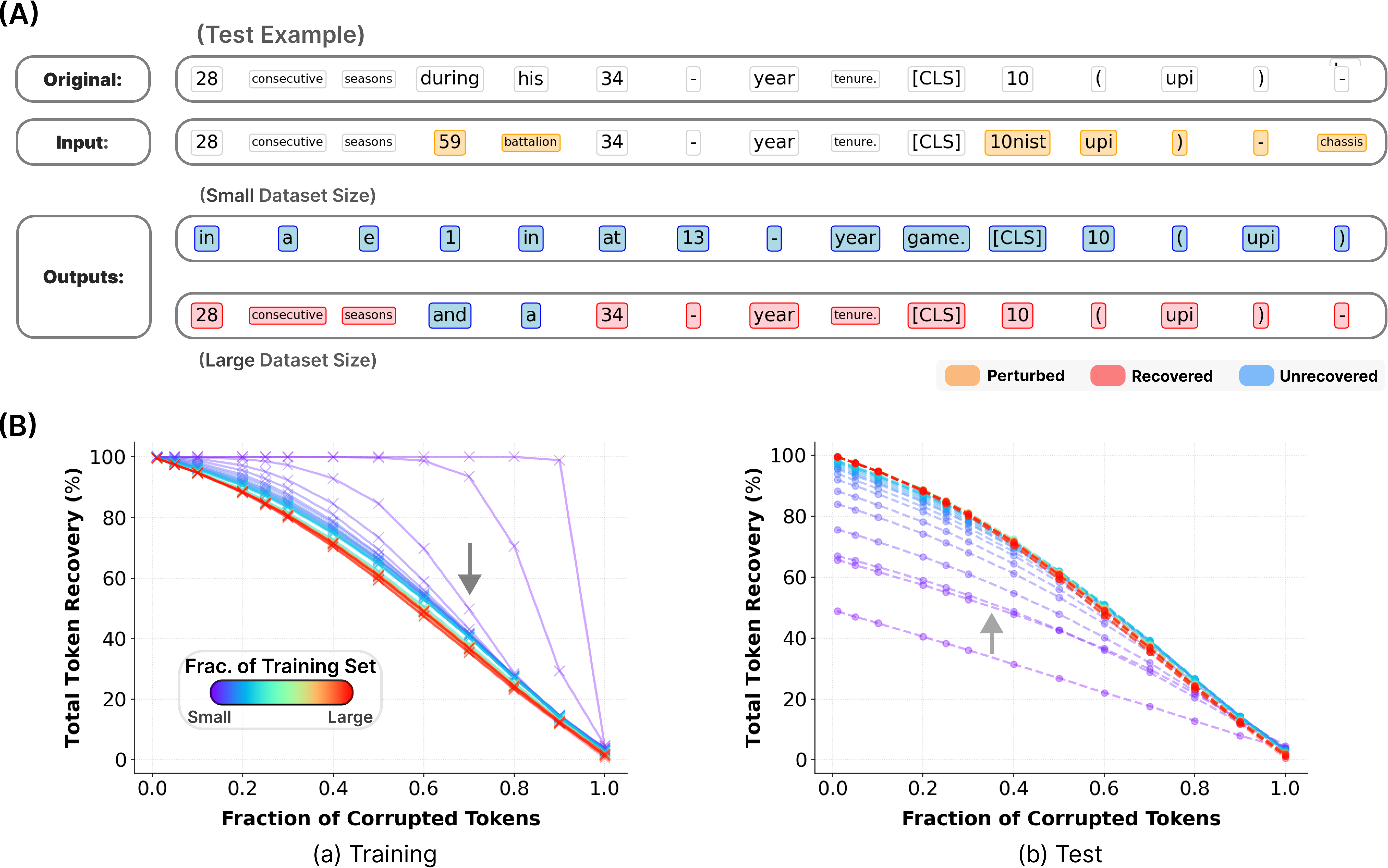}
     \vspace{-2mm}
     \caption{\textbf{Basins around training examples shrink and basins around test examples expand as the training dataset size increases}.%
     ~\textbf{(A)} Textual examples showing two \textit{Tiny} ($\sim 24$\textbf{M}) UDDMs' token recovery at noise level $t = 0.2$, where each is trained on two different training dataset sizes. With a small training dataset, the model fails to recognize unseen test sequences and alters all tokens. With a larger training set, the tokens of unseen sequences however become stable and recoverable by the sampling process.~\textbf{(B)} Average total token recovery rates (\%), including both non-corrupt and corrupted tokens, for training and test sequences across varying corruption levels. Line colors indicate the fractions of the training dataset used (ranging from \textcolor[rgb]{0.4, 0.0, 0.6}{small} to \textcolor{red}{large}). As data scales, the model’s ability to flawlessly recover training examples drops (indicating shrinking basins); while its recovery rate of unseen test examples improves (indicating expanding basins). The convergence of these rates at large training set sizes (red curves) marks the transition from memorization to generalization.
     \textit{Note:~Deterministic (greedy) sampling was used across these experiments to isolate from stochastic noise.}
     }
    \label{fig:basin_shrinkage}
\end{figure*}

%% file: EMNLP/figures/recovery_rate.tex
\begin{figure*}[t]
    \centering
    \includegraphics[width=1.0\linewidth]{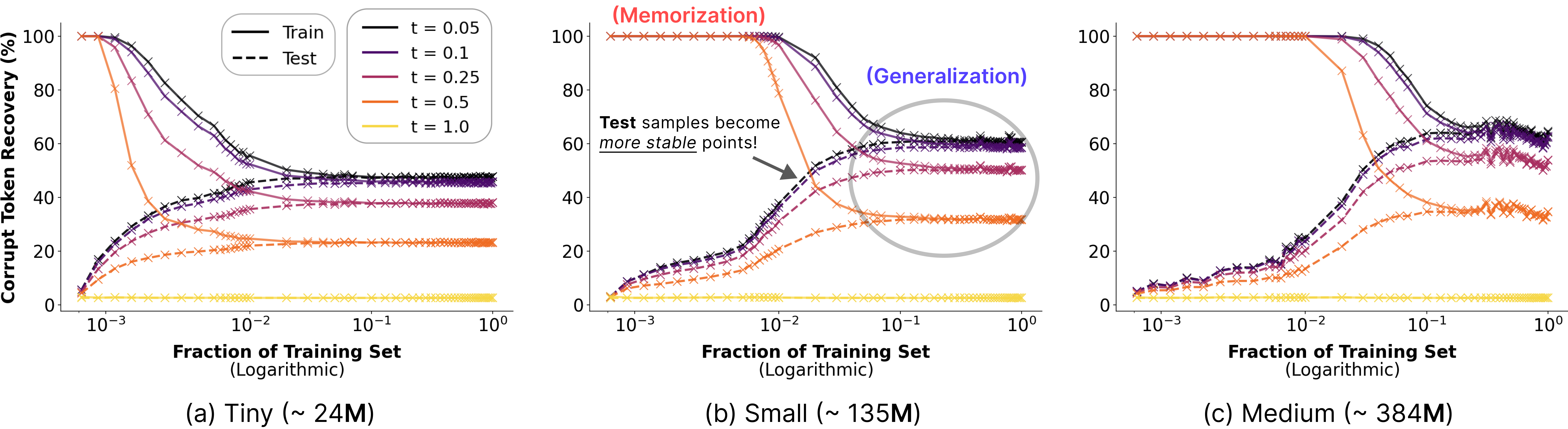}
    %\vspace{-7mm}    
    \caption{\textbf{The convergence of corrupted training and test token recovery rates marks the phase transition from memorization to generalization}. The plots display the average token recovery rate (\%) as the fraction of the training dataset grows, comparing the model's ability to denoise perturbed tokens (using \textit{stochastic sampling}) from the training set (\textbf{solid} lines) against unseen test sequences (\textbf{dashed} lines) across various perturbation levels based on time $t$ on the LM1B dataset \citep{chelba2013one}. In the limited-data regime, the model exhibits memorization, perfectly recovering training samples while failing on test data. As the dataset expands, the recovery rates for training and test sequences converge to an identical recovery rate, demonstrating that unseen samples have become stable attractors. Notably, the model's size dictates the timing of this shift, where the \textit{Medium} model requires a significantly larger fraction of data to trigger the transition, effectively prolonging the memorization phase.}
    \label{fig:token_recovery_rate_combined}
\end{figure*}

%% file: EMNLP/figures/unrecovered_recovered.tex
\begin{figure*}[t]
    \centering
    \includegraphics[width=1.0\linewidth]{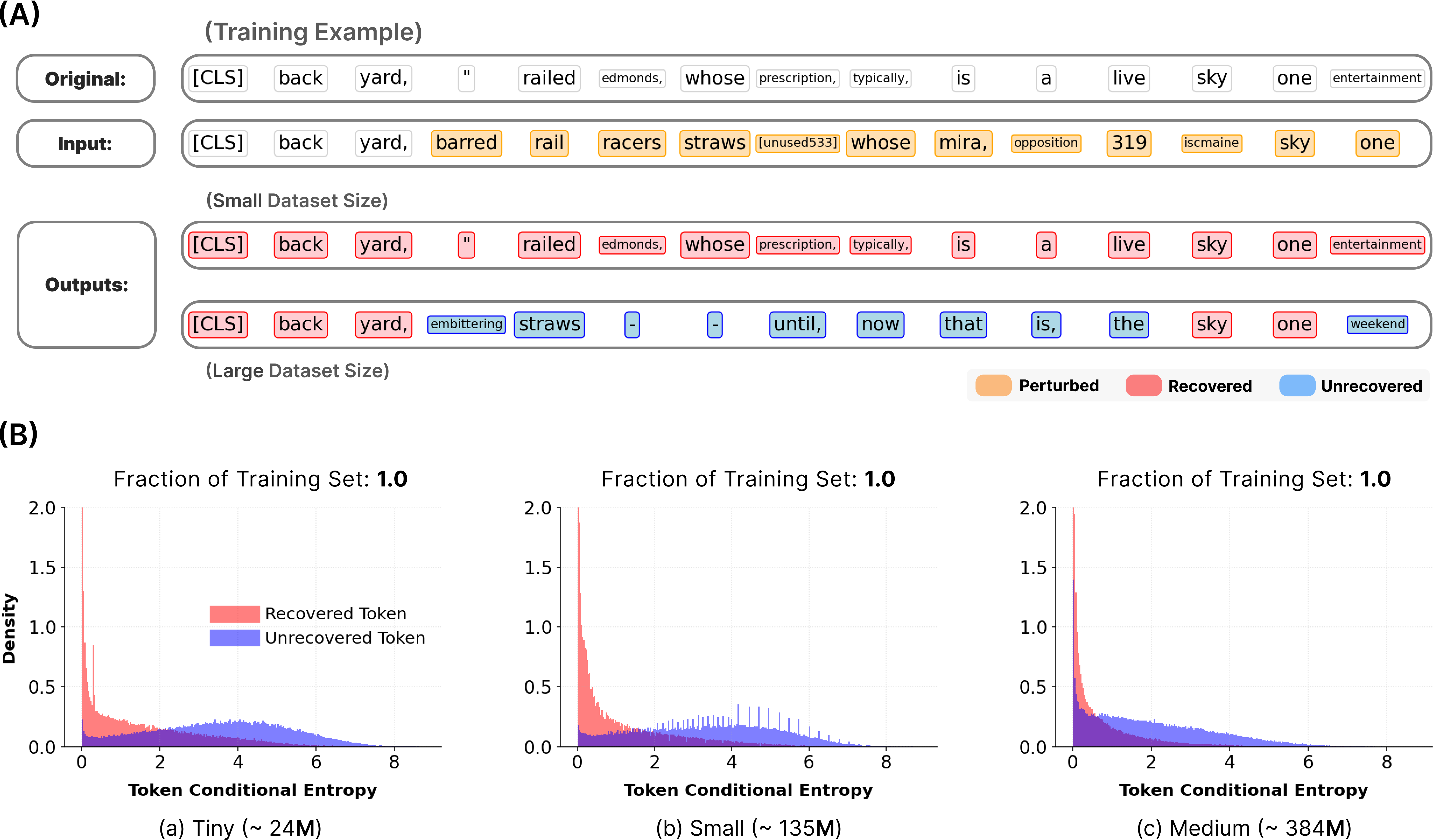}
    \vspace{-2mm}
    \caption{
    \textbf{Token-level conditional entropy highlights different token recovery behaviors}.~\textbf{(A)} Textual examples showing how two \textit{Medium} UDDMs, trained on two different training dataset sizes, recover corrupted training sequences at noise level $t = 0.5$. The model trained on a small training set perfectly memorizes and restores the original text. In contrast, the model trained on a large training set recovers some original words but actively alters others to synthesize a novel sentence.~\textbf{(B)} Density histograms of conditional entropy for individual tokens from \cref{fig:token_recovery_rate_combined}. Successfully recovered tokens consistently demonstrate much lower entropy, indicating higher model confidence. While this gap between recovered and unrecovered tokens narrows in larger models, a surprising number of highly stable, low-entropy tokens persist even during the generalization phase. \textit{Note: The $y$-axis is clipped to better contrast these two token types}.
    }
    \label{fig:unrecovered_recovered}
\end{figure*}

%% file: EMNLP/latex/section3.tex
\section{Associative Memories from Conditional Sampling}
\label{sec:am_pseudo}

%\textbf{Conditional sampling in UDDMs}. 
To establish a connection between AMs and UDDMs, we rely on a specific aspect that emerges from modeling sequences of discrete variables in DMs: the \textit{core assumption} %\citep{austin2021structured,hoogeboom2021argmax,lou2024maskeddiffusion, sahoo2024simple} 
is that the denoising process of a sequence $\rvz^{1:L}$ of length $L$ factorizes for each token $\rvz^\ell$ as 
\begin{equation}
    p^\theta_{s|t}(\rvz_s^{1:L} \mid \rvz^{1:L}_t) = \prod _{\ell=1}^L \psi^{\theta}_{s|t}(\rvz_s^\ell \mid \rvz^{1:L}_t), 
\end{equation}
where $\psi^\theta_{s|t}$ denotes the conditional probability:
\begin{equation}
%\text{Cat}\big(\rvz^\ell_s;\,\text{softmax}_K \big[ \beta(t) \, f^\ell_\theta(\rvz^{1:L}_t) \big] \big),
\psi^\theta_{s|t} (\rvz^\ell_s \mid \rvz^{1:L}_t) = \text{Cat}\big(\rvz^\ell_s;\,f^\ell_\theta(\rvz^{1:L}_t) \big).
\label{eq:categorical_conditional_sampling}
\end{equation}
%Here, $f_\theta^\ell (\mathbf{z}^{1:L}_t) = \text{softmax}_K \big[ \beta(t) \mathbf{u}^\ell(\mathbf{z}_t^{1:L}) \big ]$ , where $\mathbf{u}^\ell(\mathbf{z}_t^{1:L})$ represents the raw logits at token position $\ell$ produced by the diffusion transformer backbone , and $\beta(t)$ is a time-dependent inverse temperature. The term $\beta(t)$ is a time-dependent inverse temperature, dependent on the diffusion variable $\alpha(t)$, typically increasing as $t\to0$. These conditional probabilities enter the cross-entropy terms of NELBO (\ref{eqn:nelbo}) and are also used in practice during the denoising process. This reliance on cross-entropy is what provides the connection to AMs: as shown below and by \cite{d2025pseudo}, the cross-entropy loss produces \textit{basins of attraction around the training data points} in the dynamics of conditional sampling for a basic AM, and translates with minimal changes to UDDMs.
Here, $f_\theta^\ell (\mathbf{z}^{1:L}_t) = \mathrm{softmax}_K \big[ \beta(t) \mathbf{u}_\theta^\ell(\mathbf{z}_t^{1:L}) \big ]$, where $\mathbf{u}^\ell(\mathbf{z}_t^{1:L})$ maps the sequence $\rvz_t^{1:L}$ to raw logits via a diffusion transformer \citep{peebles2023scalable}, and the softmax is applied over the $K$-categories per position $\ell$. The inverse temperature $\beta(t)$ depends on the diffusion schedule $\alpha(t)$ and typically increases as $t\to0$. These conditional probabilities enter the cross-entropy terms in NELBO (\ref{eqn:nelbo}). This reliance on cross-entropy establishes the connection to AMs, where minimizing this loss implicitly creates basins of attraction around data points during conditional sampling \citep{d2025pseudo}, and translates with minimal changes to UDDMs.

%being the logits produced from a diffusion transformer \citep{vaswani2017attention, peebles2023scalable} and the softmax is applied over the $K$-categories, which produces a probability distribution per position $\ell$. Here, $\beta(t)$ is a time-dependent inverse temperature, dependent on the diffusion variable $\alpha(t)$, typically increasing as $t\to0$. These conditional probabilities enter the cross-entropy terms of NELBO (\ref{eqn:nelbo}) and are also used in practice during the denoising process. This reliance on cross-entropy is what provides the connection to AMs: as shown below and by \cite{d2025pseudo}, the cross-entropy loss produces \textit{basins of attraction around the training data points} in the dynamics of conditional sampling for a basic AM, and translates with minimal changes to UDDMs.

%\input{Preprint/preprint-figures/unrecovered_recovered}

\subsection{Simpler Setting: Binary Variables and Linear Logits}
%\textbf{Associative Memory using classification margins}. 
Consider an AM of $L$ binary neurons $\rvs^\ell \in \{\pm 1\}$ with a non-symmetric coupling matrix $\mathbf{W} \in \mathbb{R}^{L \times L}$, where its \textit{diagonal entries are zero}, we have the following deterministic update rule: 
\begin{equation}
\rvs^\ell_\tau  = \mathrm{sgn} \bigg ( \sum^L_{m = 1} \mathbf{W}^{\ell m } \; \rvs^m_{\tau + 1} \bigg ),
\label{eq:deterministic_update}
\end{equation}
where we adopted the notation of the discrete time $\tau$ running backwards to highlight the connection with the backward dynamics of DMs. 
Consider also a set of $P$ examples $\boldsymbol{\Xi} \in \{\pm 1\}^{P \times L}$, where $\rvx^{1:L} \in \boldsymbol{\Xi}$ is a binary vector of length $L$.

To build an AM, it is not sufficient to find an optimal coupling matrix $\mathbf{W}^*$ such that $\forall \rvx^{1:L} \in \boldsymbol{\Xi}$ is a fixed point of \cref{eq:deterministic_update}, because it would not guarantee finite basins of attraction around the examples. To create such basins, a stronger condition is instead needed:
\begin{equation}
\rvx^\ell = \mathrm{sgn} \bigg ( \sum^L_{m = 1} \mathbf{W}^{\ell m} \rvx^m + \kappa \bigg ), \quad \, \forall \ell = 1, \dots, L
\label{eq:perceptrons_margin}
\end{equation}
where the classification margin $\kappa \in \mathbb{R}^+$ ensures that each fixed point is robust to a finite amount of variable flips from the deterministic update rule \citep{gardner1988space, Gardner1988OptimalSP,forrest1988content,benedetti2022supervised}. Larger $\kappa$ implies larger basins. Given the load $\gamma=P/L$, there exists a maximum margin $\kappa_{\mathrm{max}}(\gamma)$. As shown by \citet{Soudry2018} and \citet{Montanari2024NegativePerceptron}, training a Perceptron with the cross-entropy loss in the separable regime implicitly solves \cref{eq:perceptrons_margin} with $\kappa=\kappa_{\mathrm{max}}(\gamma)$, suggesting how to produce large basins of attraction around the training examples.

\smallskip
\noindent \textbf{Conditional sampling}. To connect the cross-entropy loss with conditional sampling, it is useful to interpret this deterministic update (\ref{eq:deterministic_update}) as:
\begin{equation}
\rvs^\ell_\tau = 
\argmax_{\rvs^\ell} 
\; \psi_{\tau | \tau + 1} \big ( \rvs^\ell_\tau  \mid \rvs_{\tau + 1}^{1:L} ;  \mathbf{W}^\ell \big )
\end{equation}
where we select the most probable state under the conditional distribution (between position $\ell$ and its \textit{neighborhood} or $1 \dots L$ positions in the binary vector excluding $\ell$\footnote{Since the diagonal entries of our coupling matrix $\mathbf{W}$ are zeroed, position $\ell$ \textit{does not attend to itself}.}):
\begin{equation}
\psi_{\tau | \tau + 1} ( \rvs^\ell_\tau \; | \; \rvs_{\tau + 1}^{1:L};  \mathbf{W}^\ell) = \frac{\exp \Big ( \rvs^\ell_\tau f^\ell_{\mathbf{W}} (\rvs_{\tau + 1}^{1:L}) \Big )}{2 \cosh \Big ( f^\ell_{\mathbf{W}} (\rvs_{\tau + 1}^{1:L} ) \Big )},
\label{eq:conditional_probability_ising}
\end{equation}
where $f^\ell_{\mathbf{W}}(\rvs_{\tau + 1}^{1 : L}) = \beta \sum_{m = 1}^L  \mathbf{W}^{\ell m} \, \rvs^m_{\tau + 1}$ with a fixed inverse temperature $\beta$\footnote{\cref{eq:conditional_probability_ising} is obtained using $\psi(\rvs^\ell | \rvs^{1:L}) \propto \exp (\beta \rvs^\ell \sum_{ m} \mathbf{W}^{\ell m} \rvs^m)$.}.
For binary variables, \cref{eq:conditional_probability_ising} yields a logistic form. However, for generic categorical variables, $f_\ell(\rvs^{1:L})$ are logits inside $\mathrm{softmax}(\cdot)$ like that of   \cref{eq:categorical_conditional_sampling}. 
% Hence, \cref{eq:conditional_probability_ising} is the finite-temperature sampling of an AM and the stochastic update rule. 

\smallskip
\noindent \textbf{Conditional-likelihood maximizes classification margins}. \cref{eq:conditional_probability_ising} suggests a way to train the model: by minimizing the following loss function, the negative logarithm of the conditional-likelihood that is also called pseudo-likelihood \cite{pseudo-likelihood}:
\begin{equation}
\begin{split}
%
%&\mathcal{L}(\mathbf{W}) = -\frac{1}{P} \sum_{\rvx \; \in \; \boldsymbol{\Xi}} \; \log \prod^L_{\ell = 1} \psi(\rvx^\ell | \rvx^{1 : L} ; \mathbf{W}^\ell) \\ 
%&= -\frac{1}{P} \sum_{\rvx \; \in \; \boldsymbol{\Xi}} \sum_{\ell = 1}^L \Big [ \rvx^\ell  f^\ell_{\mathbf{W}} (\rvx^{1:L}) - \log 2 \cosh ( f^\ell_{\mathbf{W}} (\rvx^{1 : L}) ) \Big ] . 
%
\mathcal{L}(W) &= -\frac{1}{P} \sum_{\rvx \in \Xi} \log \prod_{l=1}^{L} \psi(\rvx^{l} | \rvx^{1:L}; \mathbf{W}^{l}) \\
&= -\frac{1}{P} \sum_{\rvx \in \Xi} \sum_{l=1}^{L} \Big[\rvx^{l} f_{\mathbf{W}}^{l}(\rvx^{1:L}) \\
&\,\,\,\quad - \log 2 \cosh\big(f_{\mathbf{W}}^{l}(\rvx^{1:L})\big) \Big]
\end{split}
\label{eqn:pseudo-likelihood}
\end{equation}
It has been shown by \citet{d2025pseudo} that the above objective (\ref{eqn:pseudo-likelihood}) produces \textit{basins of attraction around the training data points} in the dynamics of the conditional sampling. 
Unlike classical AM, the couplings $\mathbf{W}$ induced by conditional likelihood do not need to be symmetric, and therefore \textit{no explicit global energy function is required}. The existence of attractor-like behavior follows directly from the structure of the conditional probabilities.

To have an intuition on why the loss (\ref{eqn:pseudo-likelihood}) promotes classification margins, we can derive it with respect to the coupling matrix $\mathbf{W}$:
%we can study $\frac{\mathrm{d} \, \mathcal{L}(\mathbf{W})}{ \mathrm{d} \, \mathbf{W}^{\ell m}}$ where it is:
\begin{equation}
    %\frac{\mathrm{d} \, \mathcal{L}(\mathbf{W})}{ \mathrm{d} \, \mathbf{W}^{\ell m}} \propto - \frac{1}{P} \; \sum_{\rvx \; \in \; \boldsymbol{\Xi}} \;\;  \underset{\text{Hebbian}}{\underbrace{\rvx^\ell \rvx^m}} \Big [\underset{\text{Penalty}}{\underbrace{1 - \tanh(M^\ell (\rvx^{1:L}))}} \Big ],
\begin{split}
\frac{\dd \mathcal{L}(\mathbf{W})}{\dd \mathbf{W}^{lm}} \propto & -\frac{1}{P} \sum_{\mathbf{x}\in\Xi} \; \underbrace{\rvx^l \rvx^m}_{\text{Hebbian}} \\
& \times \Big [\underset{\text{Penalty}}{\underbrace{1 - \tanh(M^\ell (\rvx^{1:L}))}} \Big ].
\end{split}
\label{eqn:grad_pseudo}
\end{equation}
%where we highlighted the local classification margin $M^\ell(\rvx^{1:L}) = \rvx^\ell f^\ell_{\mathbf{W}}(\rvx^{1:L})$ and factored out the Hebbian term using $\tanh (f^\ell_{\mathbf{W}}(\rvx^{1:L})) = \tanh(\rvx^\ell M^\ell(\rvx^{1:L}))= \rvx^\ell \tanh (M^\ell(\rvx^{1:L}))$ and $\rvx^\ell \rvx^\ell =1$. We have two gradient terms, one that involves the typical Hebbian learning \cite{Hebb1949} used for storing data points in the Hopfield network, while the other involves modifying the classification margin around those points. The gradient penalty $1 - \tanh(M^\ell (\rvx^{1:L})) \approx 2 e^{- 2 M^\ell(\rvx^{1:L})}$ decays exponentially for correctly classified patterns with wide margins. Consequently, objective (\ref{eqn:pseudo-likelihood}) concentrates learning on patterns with the smallest margins, and it is minimized when all margins are as large as possible (the weight magnitudes diverge in the separable regime \cite{Soudry2018}).
%
Here, we denote the local classification margin as $M^\ell(\rvx^{1:L}) = \rvx^\ell f^\ell_{\mathbf{W}}(\rvx^{1:L})$. The Hebbian term is factored out via identity $\tanh (f^\ell_{\mathbf{W}}(\rvx^{1:L})) = \tanh(\rvx^\ell M^\ell(\rvx^{1:L})) = \rvx^\ell \tanh (M^\ell(\rvx^{1:L}))$, given $\rvx^\ell \rvx^\ell = 1$. This yields two distinct gradient terms: a classical Hebbian learning \cite{Hebb1949} component for pattern storage and a margin-modifying component. Because the gradient penalty $1 - \tanh(M^\ell (\rvx^{1:L})) \approx 2 e^{- 2 M^\ell(\rvx^{1:L})}$ decays exponentially for correctly classified patterns with wide margins, the objective concentrates learning on samples with the smallest margins. The loss is thus minimized when all margins are maximized, driving weight magnitudes to diverge in the separable regime \cite{Soudry2018}. 

Furthermore, we verify numerically that the optimization of loss \cref{eqn:pseudo-likelihood} leads to basins around training examples in \cref{sec:numerical_validation_of_PL} of the Appendix. Specifically, we confirm the results of \cite{Soudry2018} by reporting the lower bound computed from the idealized $\kappa_{\max}(\gamma)$ and the one computed from the empirically achieved $\kappa_{\text{achieved}}(\gamma)$ in \cref{fig:basin_verification} of the Appendix. Notably, the shrinking of basins demonstrated in \cref{fig:token_recovery_rate_combined} for UDDMs is also exhibited within the simple model trained using pseudo-likelihood (\ref{eqn:pseudo-likelihood}), supporting that the fundamental AM properties in UDDMs do not depend on the NELBO objective but on the underlying conditional-likelihood structure.

\smallskip \noindent
\textbf{Main differences with UDDMs.}~The above objective (\ref{eqn:pseudo-likelihood}) is directly related only to the term $\mathcal{L}_\text{reconstruction}$ in \cref{eqn:nelbo}, since AMs traditionally rely on a fixed temperature rather than annealing the temperature during its dynamics. By following this analogy, the reverse diffusion process can be interpreted as a stochastic AM retrieval dynamics for categorical variables where we also anneal the temperature, similarly to the continuous setting \cite{ambrogioni2023search, pham2025memorization}. From this perspective, we suspect that the additional terms in \cref{eqn:nelbo} are useful to enlarge basins of attraction when conditional probabilities are parametrized with deep architectures (like a transformer), but we leave this study for future work.

%% file: EMNLP/figures/seq_cond.tex
\begin{figure*}[t]
    \centering
    \includegraphics[width=1.\linewidth]{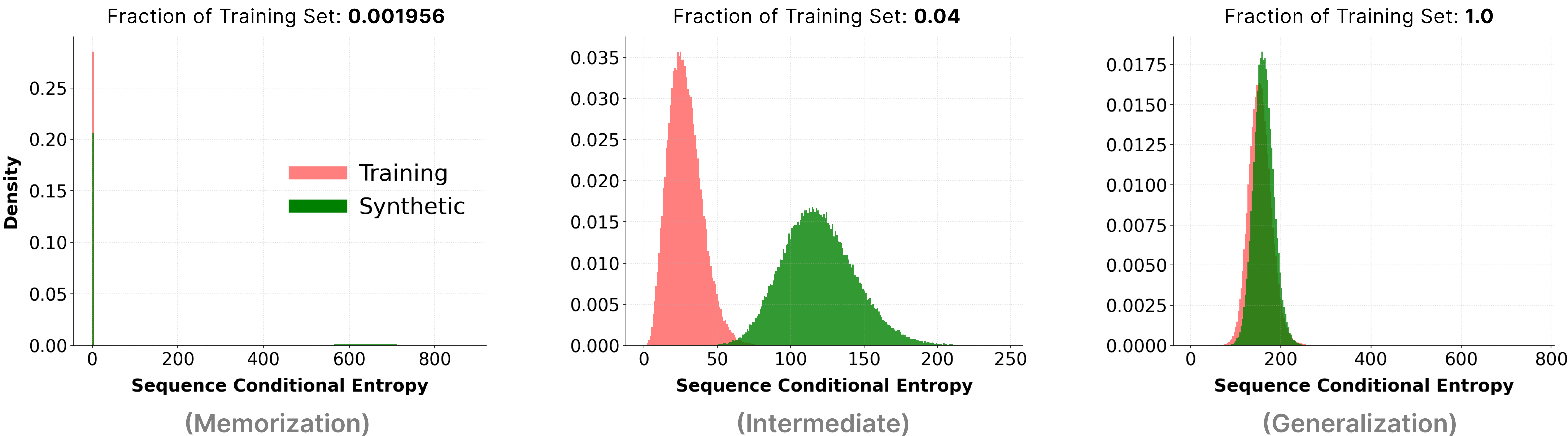}
    %\vspace{-7mm}
    \caption{\textbf{Sequence conditional entropy highlights the memorization to generalization transition}. The histograms compare the sequence entropy distributions of \textcolor{red}{training} and \textcolor[rgb]{0, 0.6, 0.2}{synthetic} (or generated) sequences respectively, for three selected fractions of the training set for the \textit{Medium} UDDMs. At a very low fraction of the training set, the conditional entropy of training sequences is near zero, aligning very well alongside the conditional entropy of many generated sequences, indicating the regime of memorization. However, as the fraction of the training set reaches the total, the synthetic and training distributions of the sequence conditional entropy converge, signifying generalization. See Appx.~(\ref{sec:add-results}) for the full range of histograms across various model sizes.
    }
    \label{fig:seq_cond}
\end{figure*}

%% file: EMNLP/figures/perplexity.tex
\begin{figure*}[t]
     \centering
     \includegraphics[width=1.\textwidth]{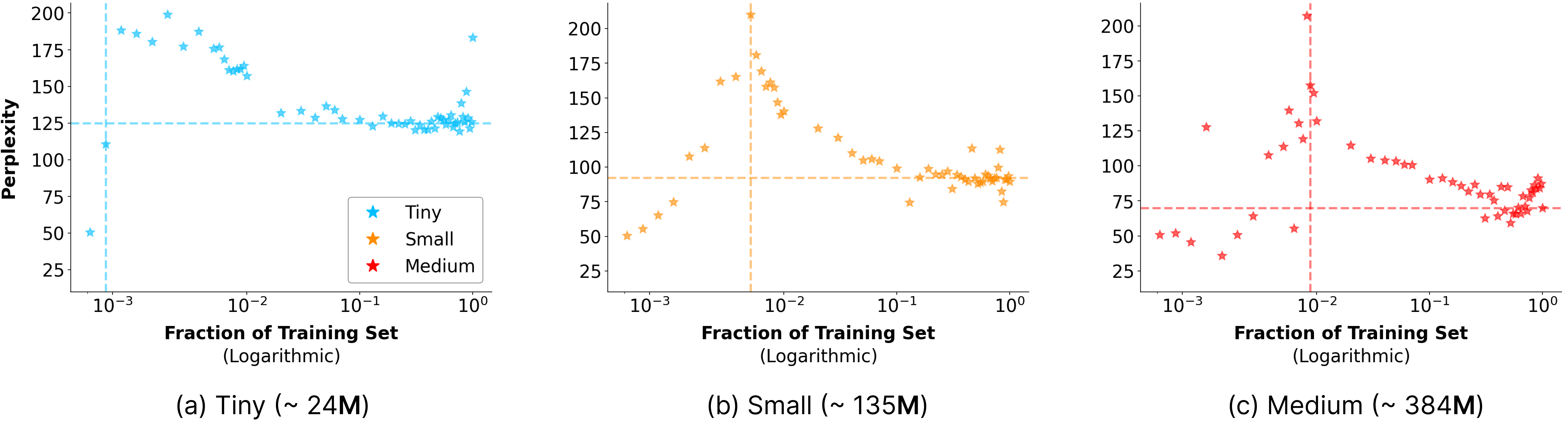}
     %\vspace{-6.5mm}
     \caption{\textbf{Perplexity trends during the memorization-to-generalization transition}. Evaluation of UDDMs across different model sizes as the fraction of the training set increases. 
     The \textit{vertical} dashed lines roughly illustrate the transition point where the model switches from memorization to generalization: perplexity starts at a very low value during the memorization regime, increases as the training dataset size grows, and it then decreases to a lower value upon reaching generalization.
     The \textit{horizontal} dashed lines highlight the converging perplexity value, showing that scaling up the model leads to improved perplexity at the end of this transition.
     %
     %Initially, during memorization, perplexity is very low. However, as the training dataset size increases, the perplexity value increases until generalization (with a sufficient number of data points), where it decreases to a smaller perplexity value. As commonly observed, increasing the model's size also leads to a better perplexity value at the end of the transition.
     %
     The precise position of each vertical line is the last point where the recovery rate is $100\%$ for training tokens in \cref{fig:token_recovery_rate_combined}. 
     \textit{Note: Perplexity is computed using GPT-2 Large \cite{radford2019language} as the critic model.}}
     \label{fig:perplexity}
\end{figure*}

%% file: EMNLP/latex/appendix_revised.tex
\section{Additional Details on Memorization to Generalization}
\label{sec:experimental-details}

\begin{table*}[t]
\centering
\begin{threeparttable}
\caption{A table showing hyperparameters of the UDDMs for the \textit{Tiny}, \textit{Small}, and \textit{Medium} sets. These variables are obtained from \cite{sahoo2025duality}.}
\label{tab:model_configs}
\begin{tabular*}{\textwidth}{@{\extracolsep\fill}lccc}
\toprule
& \multicolumn{3}{@{}c@{}}{\textbf{Model Type}} \\ \cmidrule{2-4}
\textbf{Hyperparameters} & Tiny & Small & Medium \\
\midrule
Hidden Size & 256 & 768 & 1024 \\ 
Conditioning Size & 128 & 128 & 128\\ 
Length & 128 & 128 & 128 \\ 
Num. of Blocks & 8 & 12 & 24 \\ 
Num. of Heads & 8 & 12 & 16 \\
Scale by Sigma\tnote{1} & True & True & True \\ 
Dropout & 0.1 & 0.1 & 0.1 \\
\bottomrule
\end{tabular*}
\begin{tablenotes}
    \small
    \item[1] Note: scale by sigma indicates that the model takes the inverse temperature or the appropriate diffusion scheduling parameter at time $t$ instead of the typical approach of conditioning on $t$.
\end{tablenotes}
\end{threeparttable}
\end{table*}

\noindent \textbf{Setup}. For our experiments, showcased in \cref{sec:results}, we trained three sets of UDDMs, labeled as \textit{Tiny}, \textit{Small}, and \textit{Medium}, utilizing the code base and approach of \citet{sahoo2025duality}\footnote{The code-base used in our experiments is publicly available at \href{https://github.com/s-sahoo/duo}{https://github.com/s-sahoo/duo} under the Apache 2.0 license, and we use it solely for research purposes consistent with its intended use.
}. For more details on this variant of UDDMs, please refer to the discussion in Appx.~(\ref{sec:duo}) below. Meanwhile, the backbone of our trained UDDMs is the diffusion transformer architecture from \cite{peebles2023scalable}. The configurations of our three variations of UDDMs are described in \cref{tab:model_configs} \footnote{The LM1B dataset \cite{chelba2013one} is publicly available at \href{https://www.statmt.org/lm-benchmark/}{https://www.statmt.org/lm-benchmark/} and is intended for research purposes in statistical language modeling, which is consistent with our use.}.

Meanwhile, there are a total of $162$ models, or $54$ models for each of the three UDDM sets we have trained. All models are trained up to 1 million training iterations following the procedures detailed in \citet{sahoo2025duality}. For the selection of the fraction of the training dataset sizes, we initially start with the fraction $n = 10^{-2}$ and increment it by $ \Delta n = 0.03$ all the way to the full dataset. However, to further magnify the memorization phase, we train more points using linear spacing (of $17$ points, inclusively) starting at $10^{-4}$ to $10^{-2}$, and another set of points using linear spacing (of $7$ points, inclusively) starting from $10^{-2}$ to $0.07$. Lastly, our models are trained on the LM1B dataset \citep{chelba2013one}, using \texttt{GPT-2} tokenizer \citep{radford2019language}, where our model handles the block size (or sequence length) of $128$, and all of them are initialized from the same random seed.

\smallskip
\noindent \textbf{Token Recovery Rate}. To obtain the results in \cref{fig:token_recovery_rate_combined}, we utilized our trained models, from the three sets corresponding to \textit{Tiny}, \textit{Small}, and \textit{Medium} UDDMs. Here, we computed the analysis of randomly chosen $5 \times 10^{3}$ samples belonging to their respective training set (in accordance with their fraction of training dataset size) and also $5 \times 10^{3}$ sequences from the test set. We performed the perturbation of these sequences using the forward process (\ref{eqn:forward-duo}) at time $t$, and run the reverse process (\ref{eqn:discrete-reverse-process}) starting at or near the same time $t$ back to a small terminal time $\epsilon = 10^{-5}$. Finally, to measure our recovery rate, we applied \cref{eqn:token-recovery-rate} from the main text, which measures the rate of perturbed tokens being recovered. 

Meanwhile, for \cref{fig:basin_shrinkage}, we are interested in the shrinkage of the basins of our training and unseen test sequences. To obtain a more comprehensive view of this shrinkage without worrying about the stochastic noise added during each reverse process (\ref{eqn:discrete-reverse-process}) step as $t \to 0$, we utilized the greedy or deterministic sampling dynamics instead. Specifically, in \cref{eqn:discrete-reverse-process}, to obtain $\rvz_s$ as we traverse back in time, we are effectively sampling from a categorical distribution $\text{Cat}(\cdot)$. To convert this stochastic process to a deterministic one, we simply replaced $\text{Cat}(\cdot)$ with the $\argmax (\cdot)$ operation. Similarly to the previous experiment, we applied perturbation using the forward process (\ref{eqn:forward-duo}), and utilized the greedy process to denoise our sequences. The timesteps we used are $\{0.1, 0.2, 0.3, \dots, 1.0\}$ and the same values of $t$ we used in \cref{fig:token_recovery_rate_combined}.

\begin{figure}[H]
    \centering
    \includegraphics[width=1\linewidth]{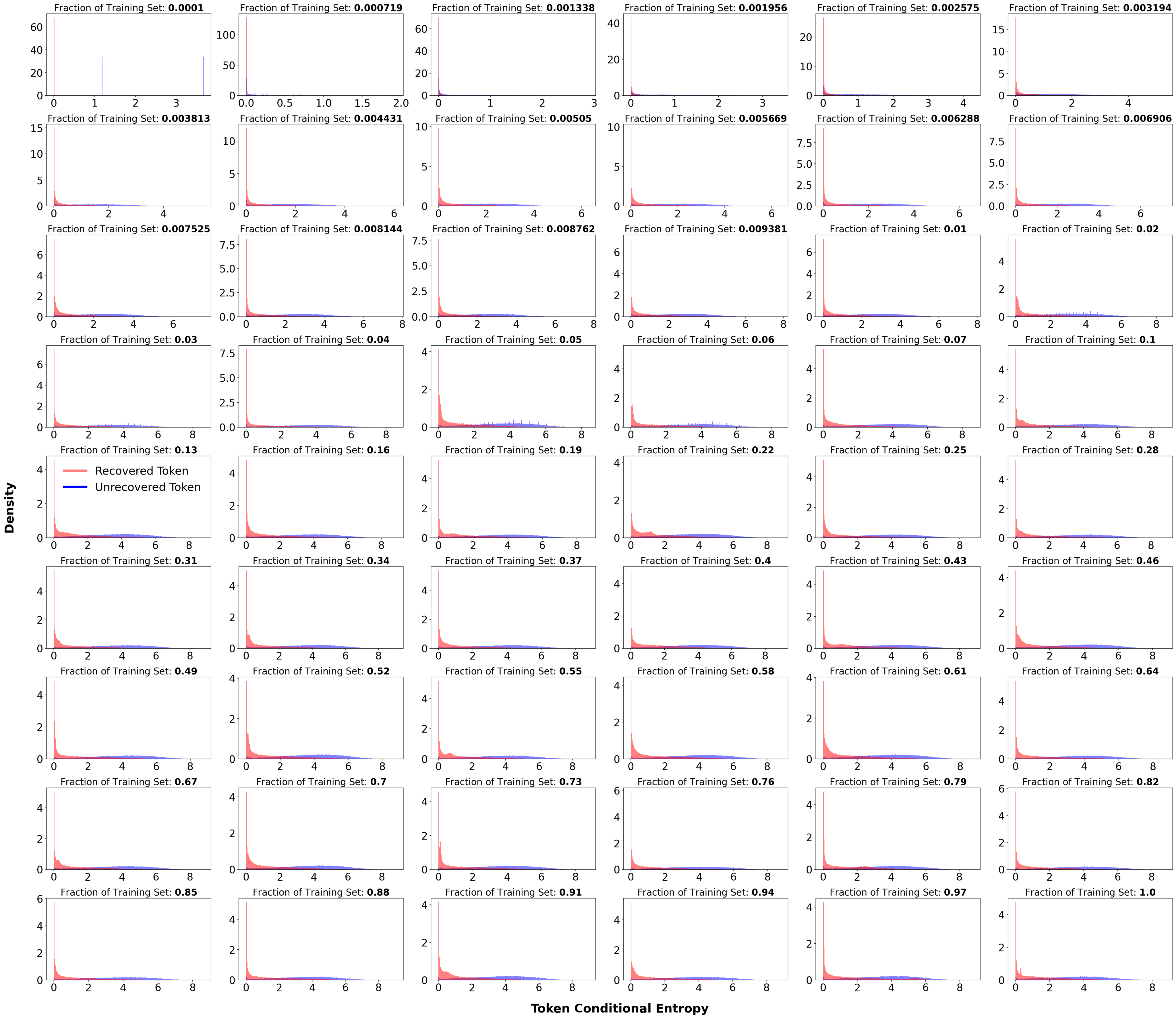}
    \caption{An illustration of the density of conditional entropy for two categories of tokens, \textcolor{red}{\textit{recovered}} and \textcolor{blue}{\textit{unrecovered}}, computed at $t = 0.25$ for the \textit{Tiny} model. The subplots are ordered by the fraction of training dataset, ranging from $10^{-4}$ (top-left) to $1.0$ (bottom-right). As the fraction of training data increases, recovered tokens concentrate near zero entropy (high confidence), while unrecovered tokens exhibit a broad distribution at higher entropy.
    }
    \label{fig:additional_recovered_unrecovered_hist_tiny}
\end{figure}

\begin{figure}[h]
    \centering
    \includegraphics[width=1.0\linewidth]{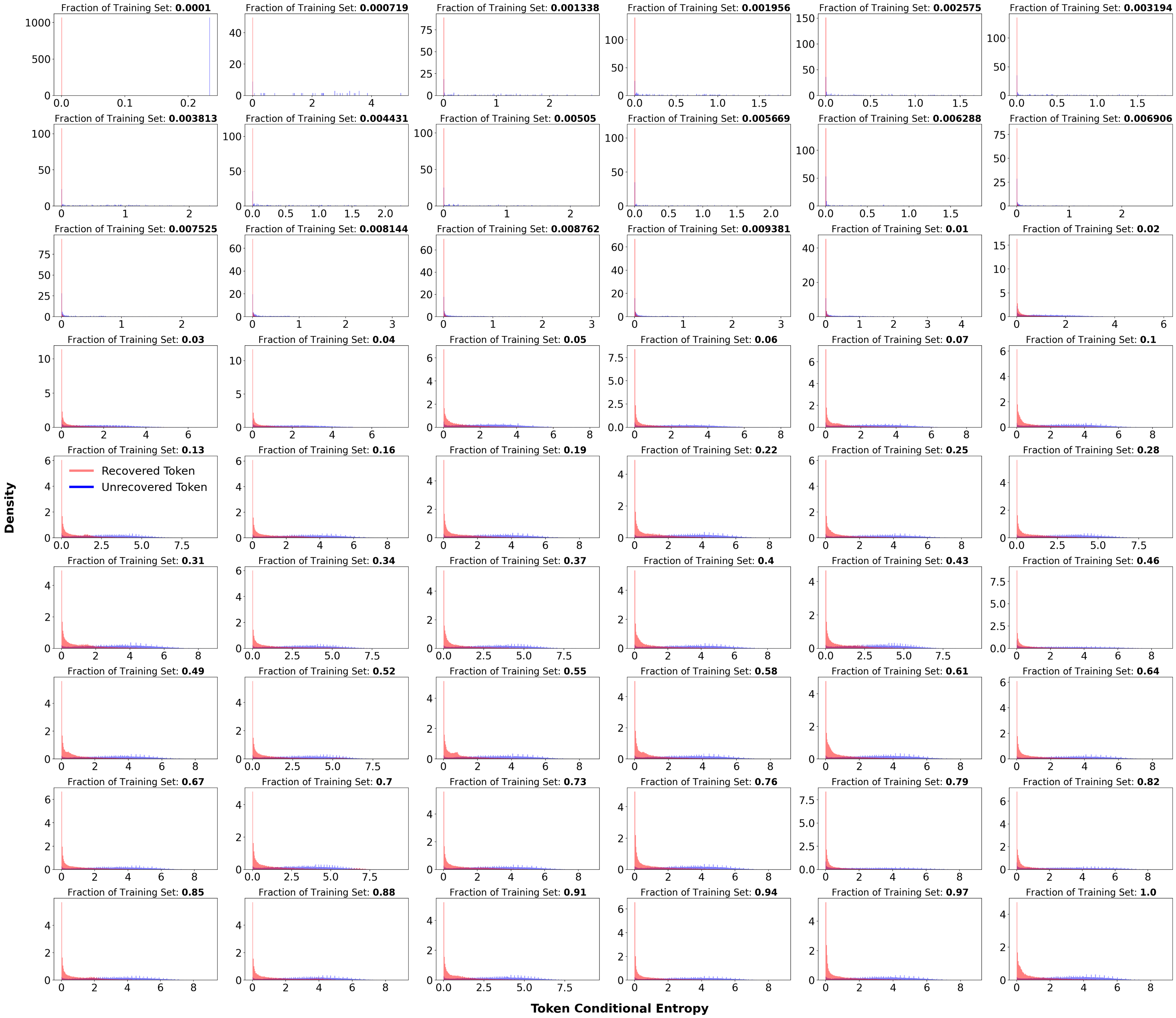}
    \caption{An illustration of the density of conditional entropy for two categories of tokens, \textcolor{red}{\textit{recovered}} and \textcolor{blue}{\textit{unrecovered}}, computed at $t = 0.25$ for the \textit{Small} model. The subplots are ordered by the fraction of training dataset, ranging from $10^{-4}$ (top-left) to $1.0$ (bottom-right). As the fraction of training data increases, recovered tokens concentrate near zero entropy (high confidence), while unrecovered tokens exhibit a broad distribution at higher entropy.
    }
    \label{fig:additional_recovered_unrecovered_hist_small}
\end{figure}

\begin{figure}[h]
    \centering
    \includegraphics[width=1.0\linewidth]{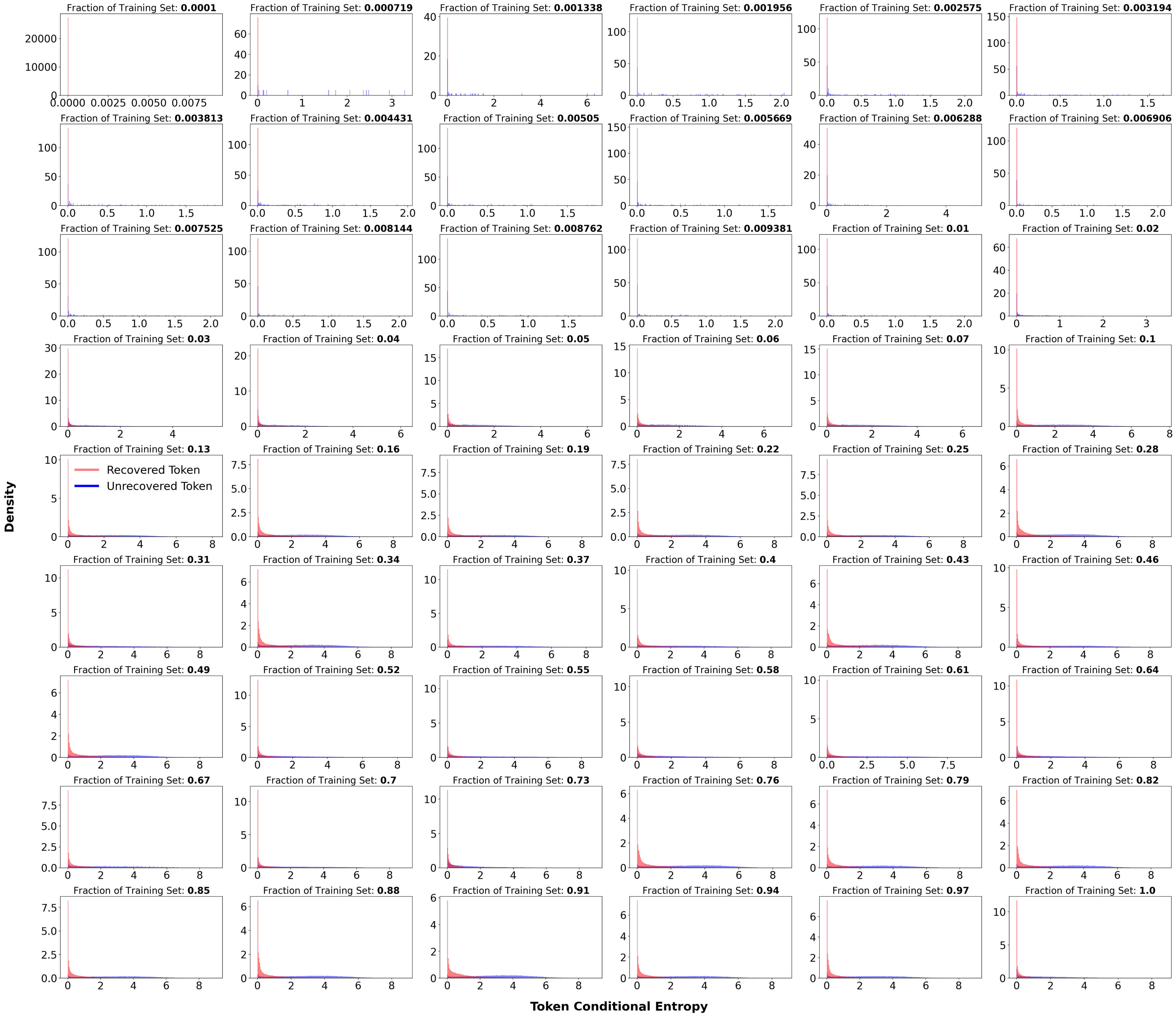}
    \caption{An illustration of the density of conditional entropy for two categories of tokens, \textcolor{red}{\textit{recovered}} and \textcolor{blue}{\textit{unrecovered}}, computed at $t = 0.25$ for the \textit{Medium} model. The subplots are ordered by the fraction of training dataset, ranging from $10^{-4}$ (top-left) to $1.0$ (bottom-right). As the fraction of training data increases, recovered tokens concentrate near zero entropy (high confidence), while unrecovered tokens exhibit a broad distribution at higher entropy.
    }
    \label{fig:additional_recovered_unrecovered_hist_medium}
\end{figure}

\begin{figure}[h]
    \centering
    \includegraphics[width=1.0\linewidth]{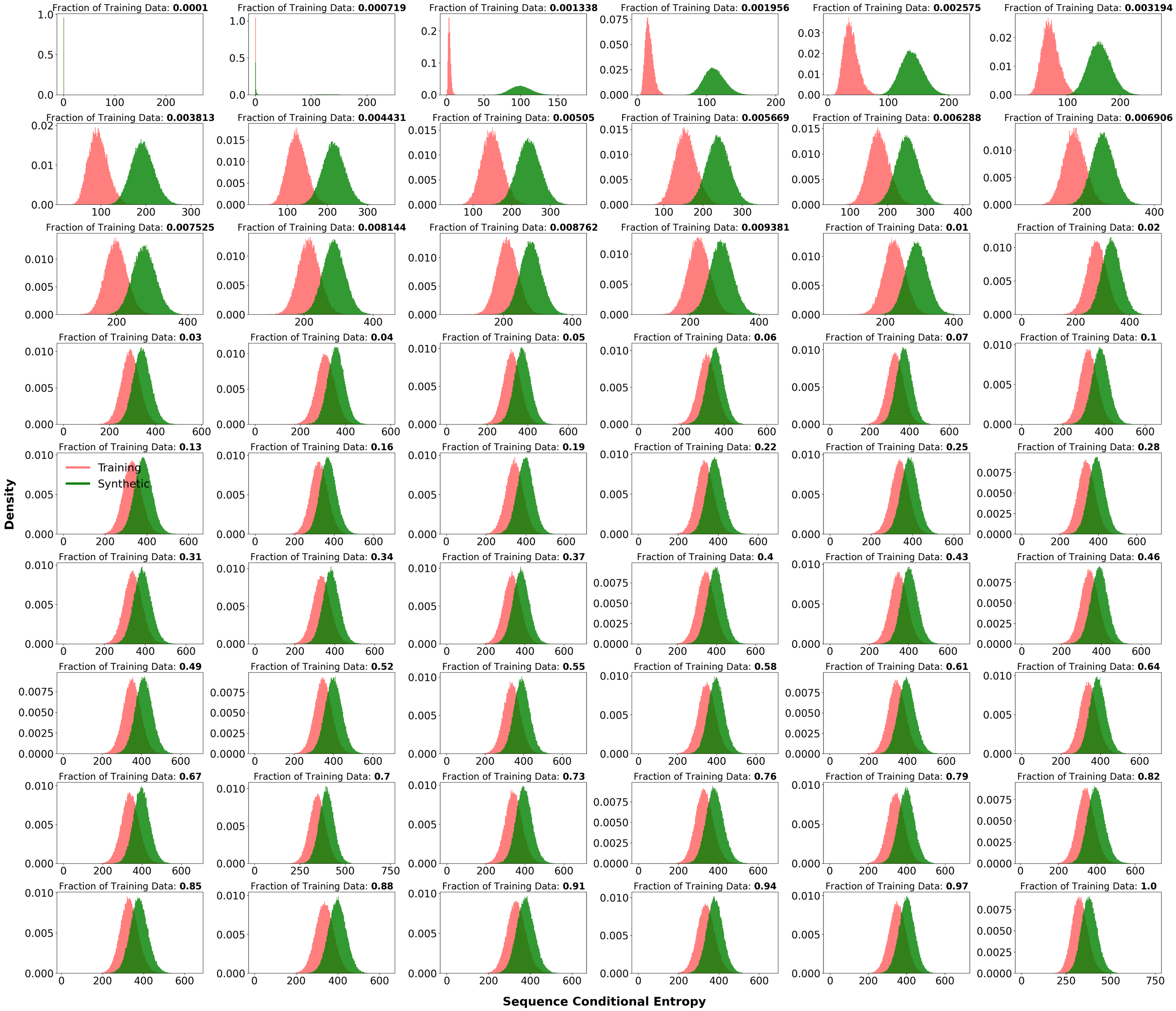}
    \caption{An illustration of the evolution of the density of the average conditional entropy for the probabilities of \textcolor{red}{training} and \textcolor[rgb]{0., 0.6, 0.2}{synthetic} sequences respectively, computed at $t = 10^{-5}$ using the \textit{Tiny} models, as the training dataset size grows. When the fraction of training set is small, there exists a separation in the average conditional entropies of training and synthetic samples. However, as the training dataset size grows, this separation is reduced and the conditional entropy of synthetic samples becomes similar to that of the training samples.
    %In the memorization phase, when the fraction of the training dataset is small, most tokens have very low conditional entropy. In contrast, during the generalization phase, many tokens have high conditional entropy, but low-entropy tokens still remain.
    }
    \label{fig:additional_train_synth_hist_tiny}
\end{figure}

\begin{figure}[h]
    \centering
    \includegraphics[width=1.0\linewidth]{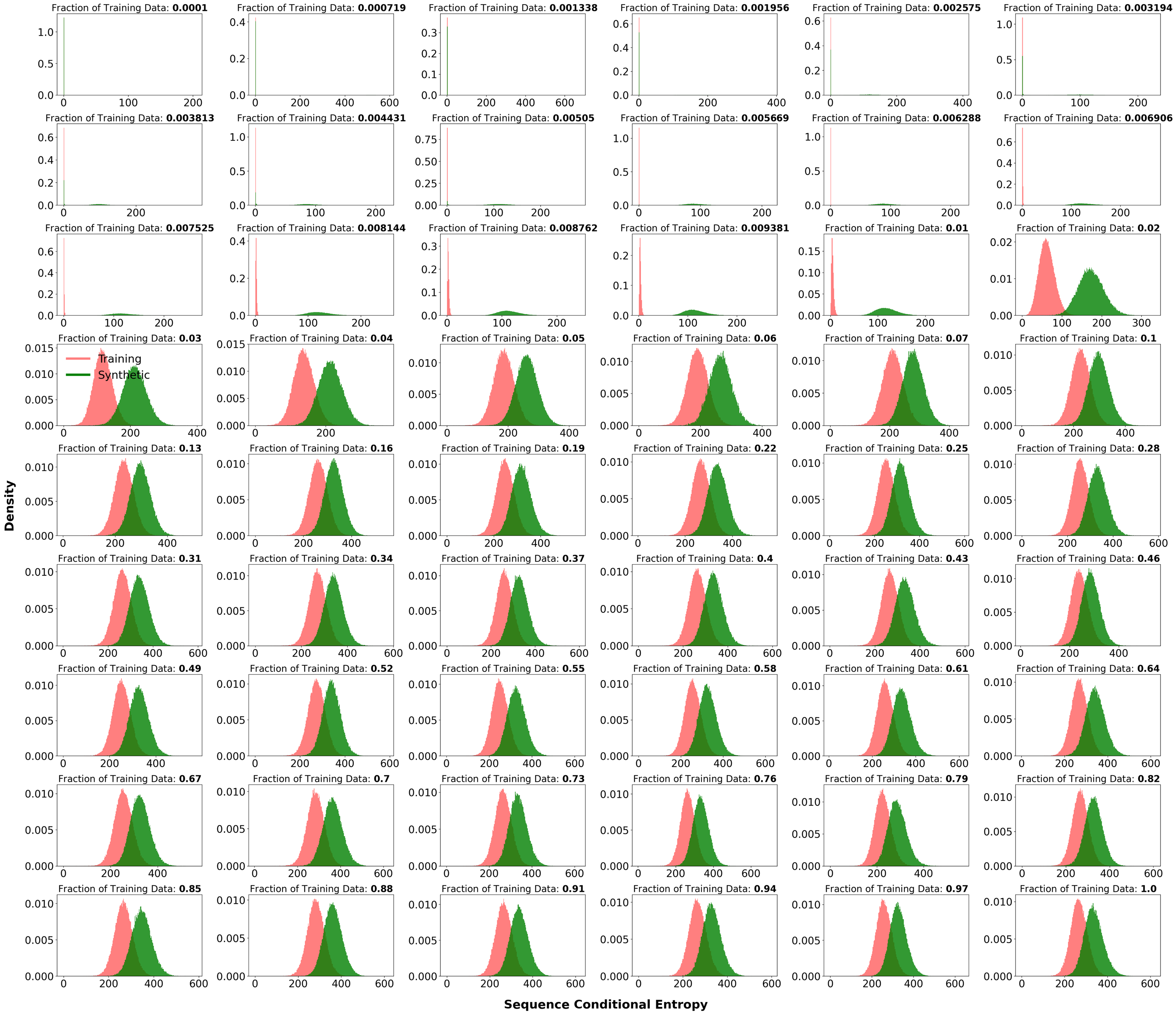}
    \caption{An illustration of the evolution of the density of the average conditional entropy for the probabilities of \textcolor{red}{training} and \textcolor[rgb]{0., 0.6, 0.2}{synthetic} sequences respectively, computed at $t = 10^{-5}$ using the \textit{Small} models, as the training dataset size grows. When the fraction of training set is small, there exists a separation in the average conditional entropies of training and synthetic samples. However, as the training dataset size grows, this separation is reduced and the conditional entropy of synthetic samples becomes similar to that of the training samples.
    %In the memorization phase, when the fraction of the training dataset is small, most tokens have very low conditional entropy, nearing zero value. In contrast, during the generalization phase, many tokens have high conditional entropy, but low-entropy tokens still remain.
    }
    \label{fig:additional_train_synth_hist_small}
\end{figure}

\begin{figure}[h]
    \centering
    \includegraphics[width=1.0\linewidth]{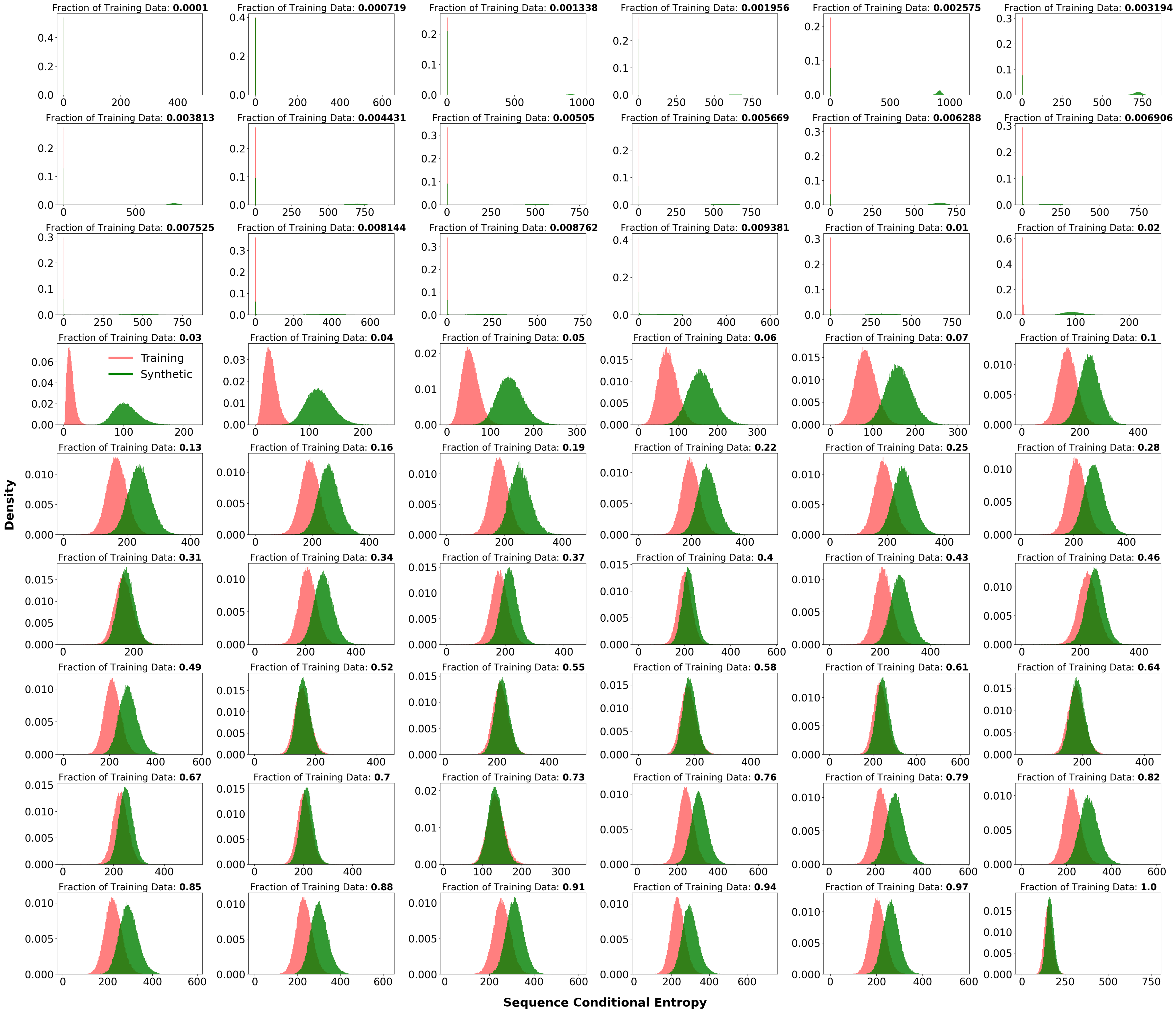}
    \caption{An illustration of the evolution of the density of the average conditional entropy for the probabilities of \textcolor{red}{training} and \textcolor[rgb]{0., 0.6, 0.2}{synthetic} sequences respectively, computed at $t = 10^{-5}$ using the \textit{Medium} models, as the training dataset size grows. When the fraction of training set is small, there exists a separation in the average conditional entropies of training and synthetic samples. However, as the training dataset size grows, this separation is reduced and the conditional entropy of synthetic samples becomes similar to that of the training samples. Due to larger model size, the UDDM exhibits stronger memorization (and chaotic) behaviors, where the two distributions initially merge at certain fractions of the training dataset and later diverge. Nonetheless, with the full training set, both distributions overlap each other at the end. 
    }
    \label{fig:additional_train_synth_hist_medium}
\end{figure}

\smallskip
\noindent \textbf{Conditional Entropy}. %Similarly, in the bottom row of \cref{fig:token_recovery_rate_combined}, we identified which tokens in a sequence is recovered and those have failed to be recovered, and then compute the conditional entropy of each class of tokens using \cref{eqn:cond-ent} in the main text. For this result, we do not average the conditional entropy. However, in the case of results in top panel of \cref{fig:entropy_over_time}, we measured the average conditional entropy of $5000$ training samples (which is also averaged over the sequence length) of the reverse process starting from $t = 1$ to $\epsilon = 10^{-5}$. Similarly, perturbation is first applied and the reverse process is performed afterwards. 
In \cref{fig:unrecovered_recovered}, we computed the token conditional entropy following \cref{eqn:cond-ent} for unrecovered and recovered tokens at $t = 10^{-5}$ identified in \cref{fig:token_recovery_rate_combined} using $5 \times 10^{3}$ samples (per training and per test sequences), respectively. Meanwhile, for the results in \cref{fig:seq_cond} and \cref{fig:entropy_over_time}, we generated a synthetic set of $10^{5}$ samples for each of the trained models and performed our analyses -- alongside using at most $10^{5}$ sequences belonging to the training set that each model was trained with. It is important to note that the results in \cref{fig:entropy_over_time} are the average conditional entropy computed across sequences and their respective tokens, while \cref{fig:seq_cond} showcases the sequence conditional entropy, which is the sum of all of the tokens' conditional entropy in a given sequence. Please see Figs.~(\ref{fig:additional_train_synth_hist_tiny})-(\ref{fig:additional_train_synth_hist_medium}) for the full histograms of conditional entropy on training versus generated samples, and Figs.~(\ref{fig:additional_recovered_unrecovered_hist_tiny})-(\ref{fig:additional_recovered_unrecovered_hist_medium}) for more conditional entropy histograms of the unrecovered versus recovered tokens shown in \cref{fig:token_recovery_rate_combined}.

\begin{figure}[htbp]
    \subsection{Shrinkage and Expansion of Basins}
    %\vspace{-7.5mm}
    \centering
    \begin{minipage}{0.925\textwidth}
        \centering
        \begin{subfigure}[b]{0.465\textwidth}
             \centering
             \includegraphics[width=\textwidth]{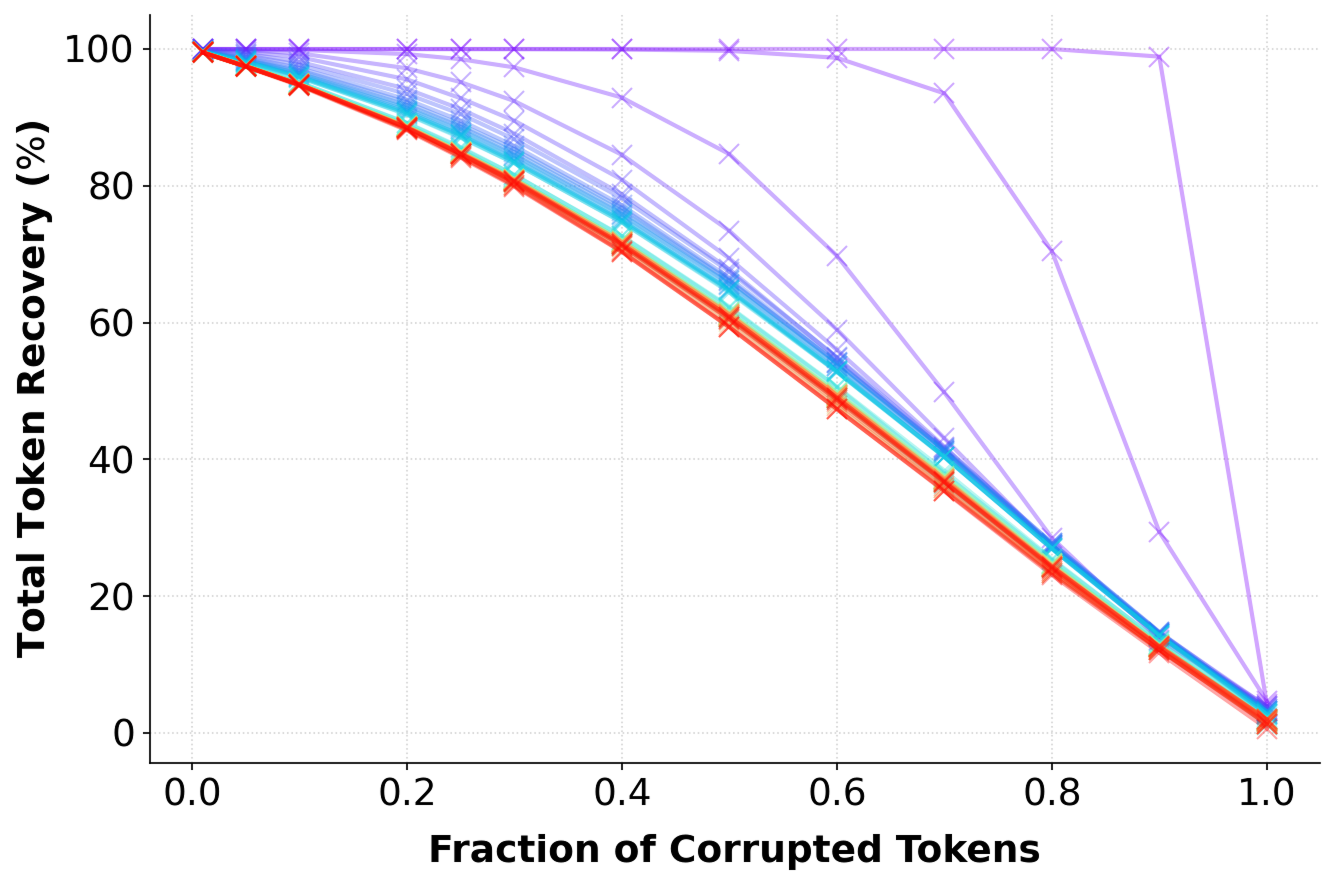}
             \caption{\fontsize{12}{16}\selectfont  Training}
        \end{subfigure}
        \hfill
        \begin{subfigure}[b]{0.465\textwidth}
             \centering
             \includegraphics[width=\textwidth]{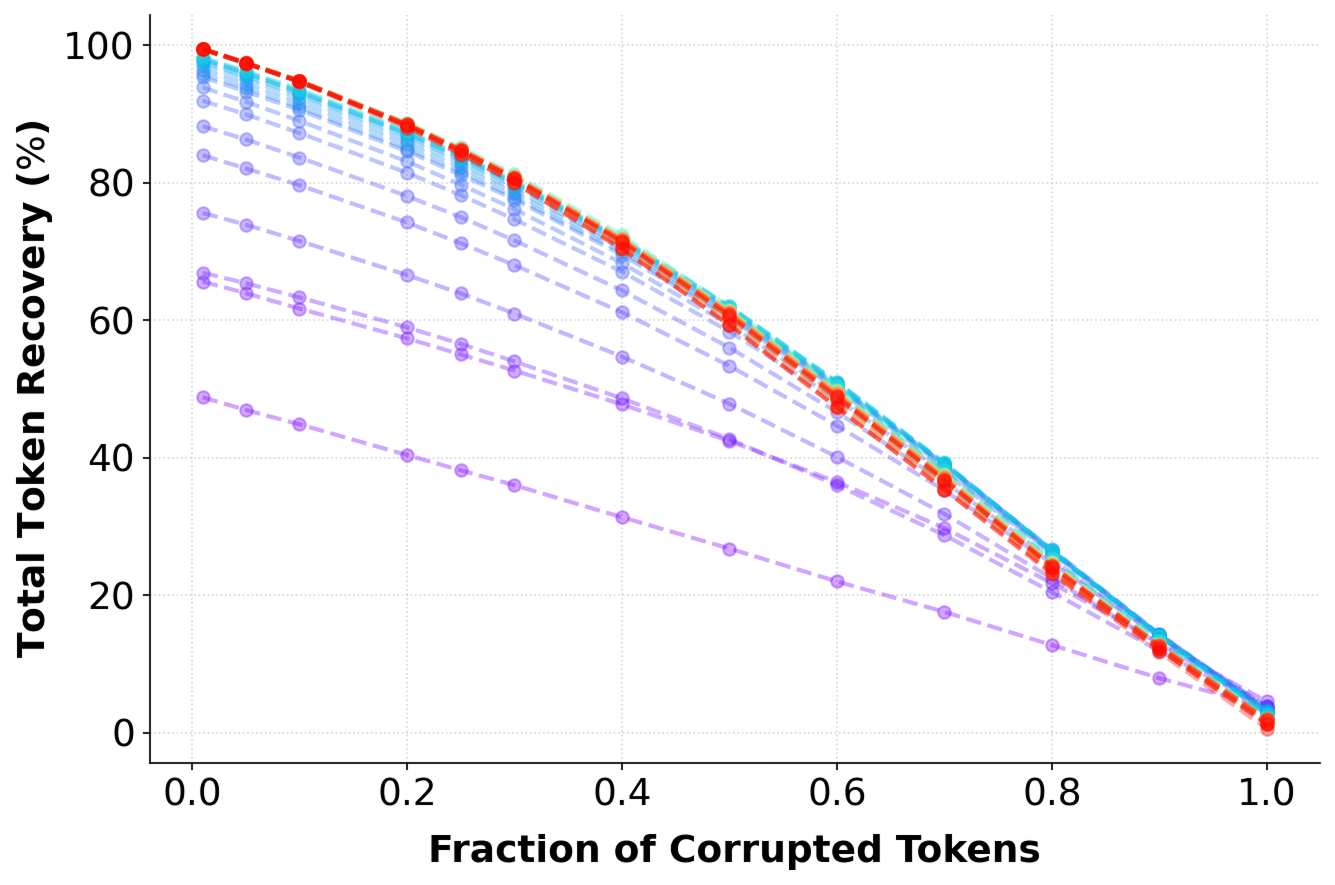}
             \caption{\fontsize{12}{16}\selectfont Test}
        \end{subfigure}
        \vspace{-3mm}
        %\captionof*{figure}{Tiny ($\sim$ 24\textbf{M})} 
        \par{Tiny ($\sim$ 24\textbf{M})}
        \vspace{3.5mm}
    \end{minipage}
    \begin{minipage}{0.925\textwidth}
        \centering
        \setcounter{subfigure}{0}
        \begin{subfigure}[b]{0.465\textwidth}
             \centering
             \includegraphics[width=\textwidth]{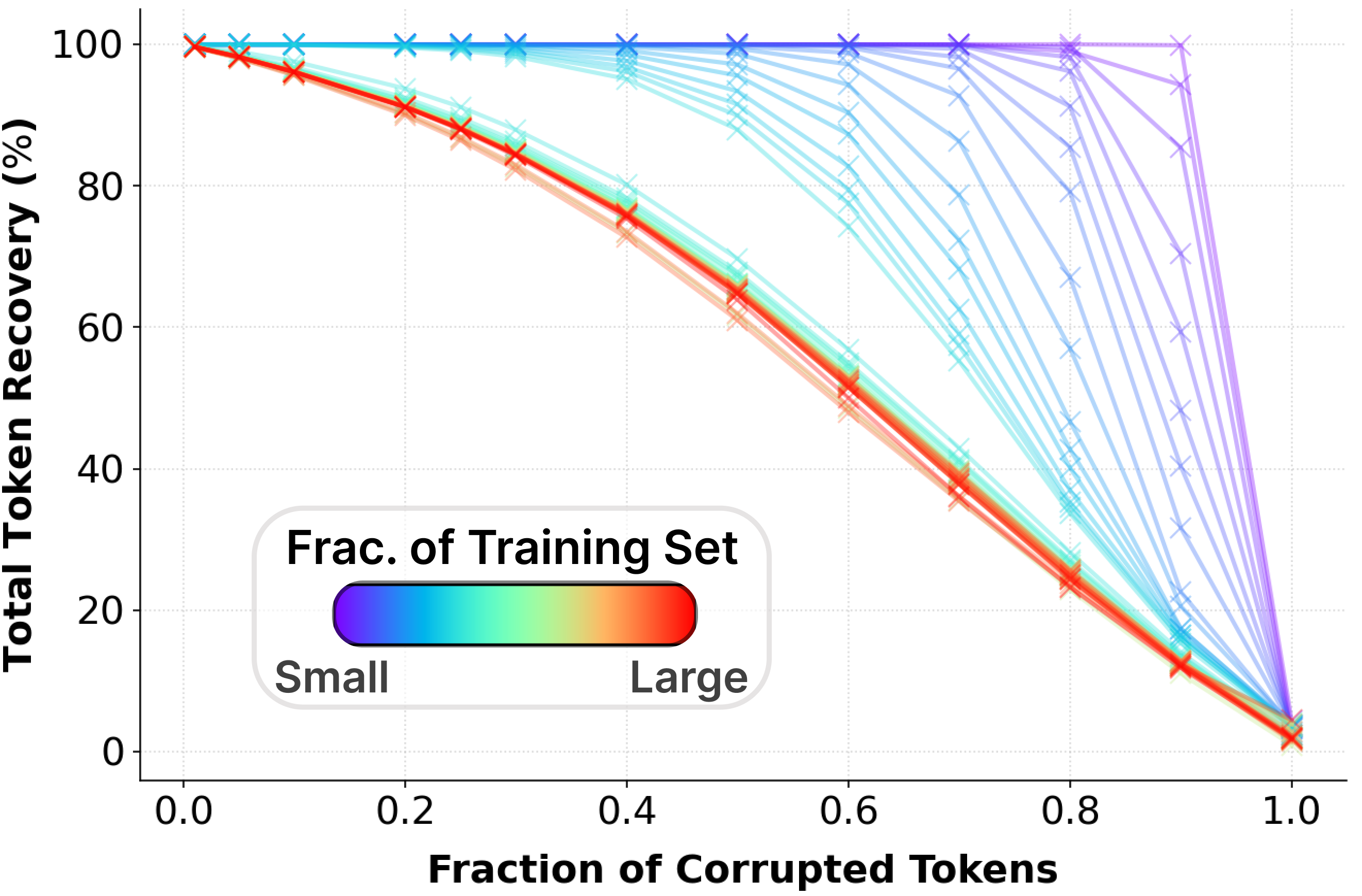}
             \caption{\fontsize{12}{16}\selectfont  Training}
        \end{subfigure}
        \hfill
        \begin{subfigure}[b]{0.465\textwidth}
             \centering
             \includegraphics[width=\textwidth]{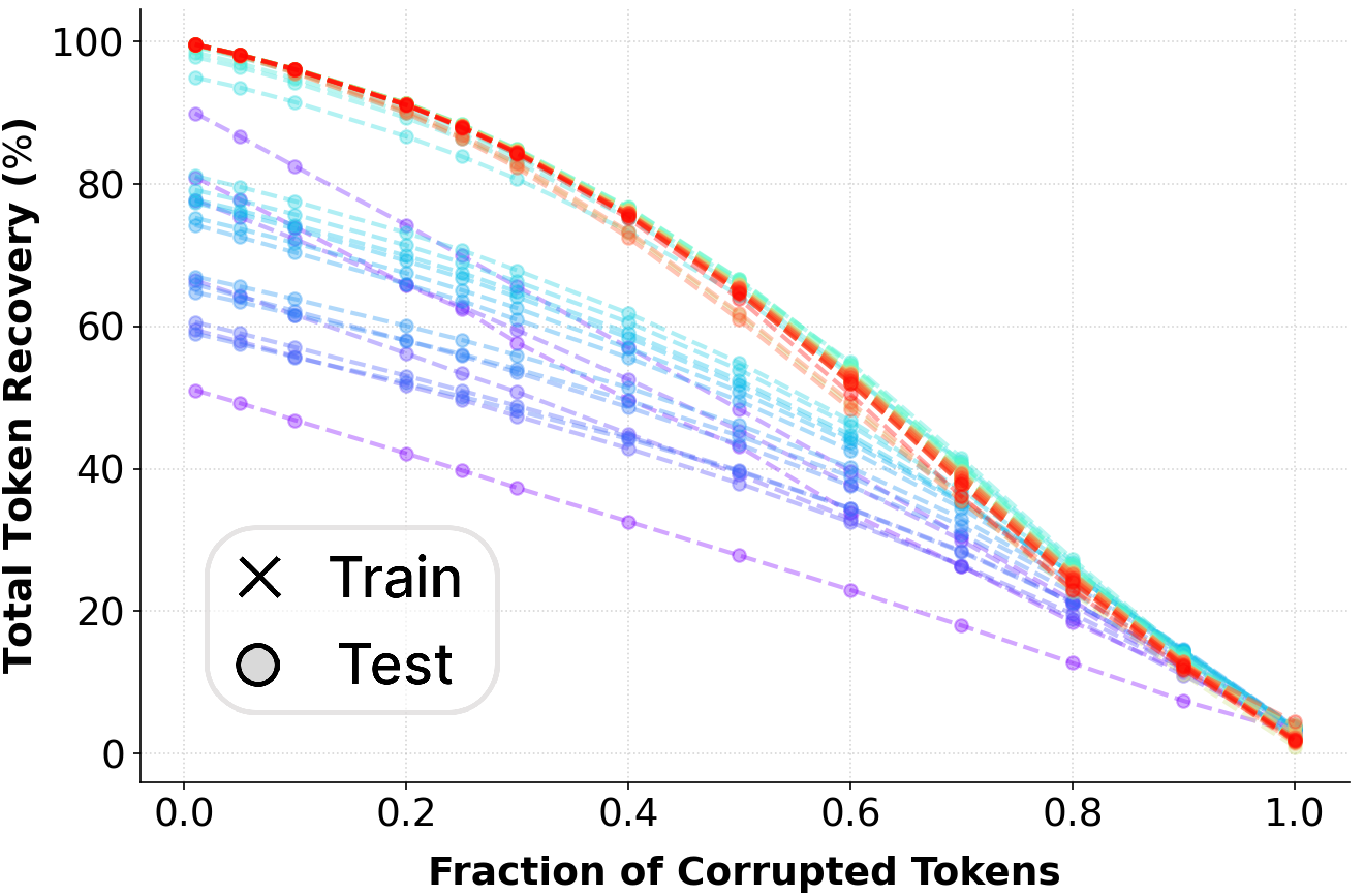}
             \caption{\fontsize{12}{16}\selectfont Test}
        \end{subfigure}
        \vspace{-3mm}
        %\captionof*{figure}{Small ($\sim$ 135\textbf{M})} 
        \par{Small ($\sim$ 135\textbf{M})}
        \vspace{3.5mm}
    \end{minipage}
    \begin{minipage}{0.925\textwidth}
        \centering
        \setcounter{subfigure}{0}
        \begin{subfigure}[b]{0.465\textwidth}
             \centering
             \includegraphics[width=\textwidth]{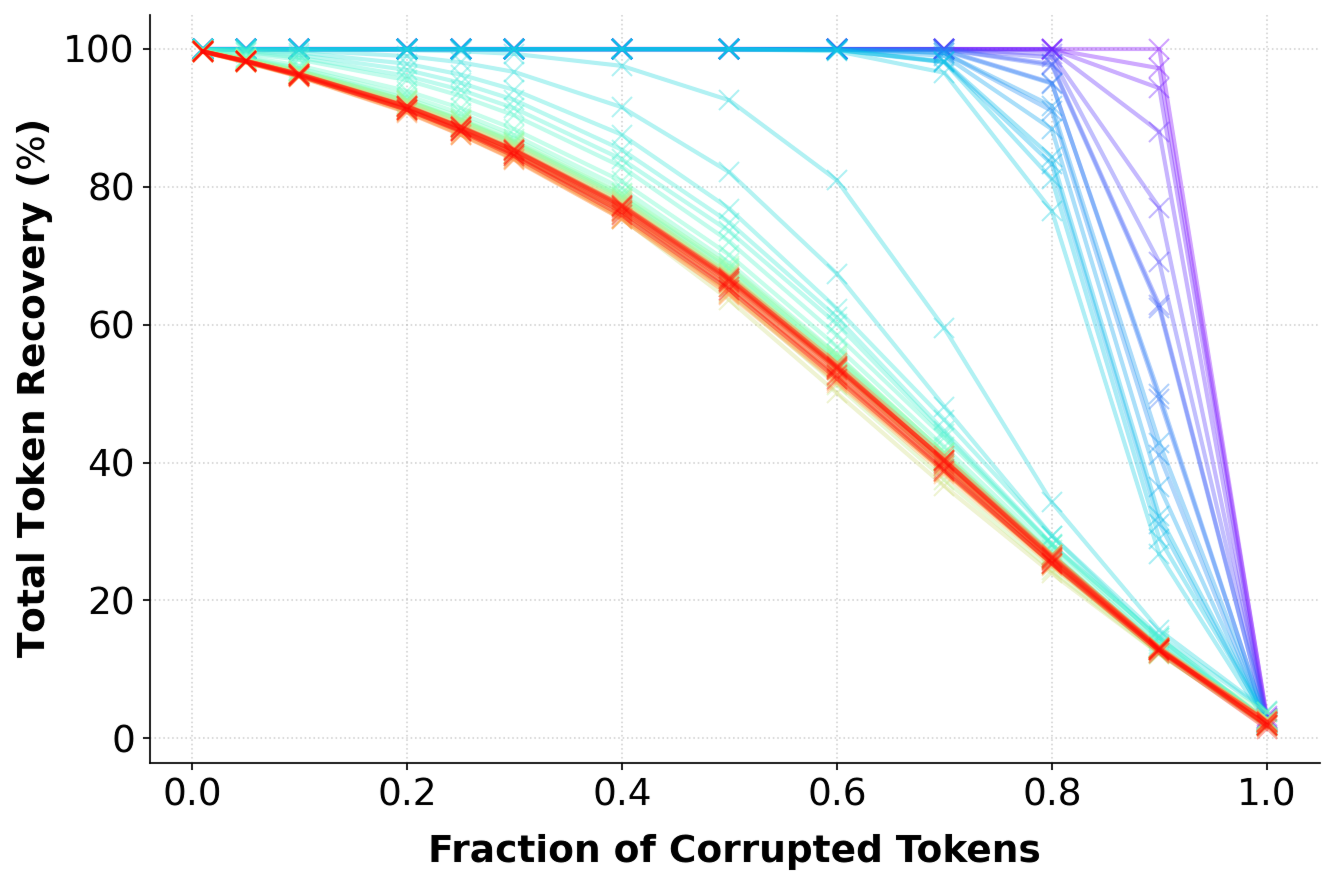}
             \caption{\fontsize{12}{16}\selectfont  Training}
        \end{subfigure}
        \hfill
        \begin{subfigure}[b]{0.465\textwidth}
             \centering
             \includegraphics[width=\textwidth]{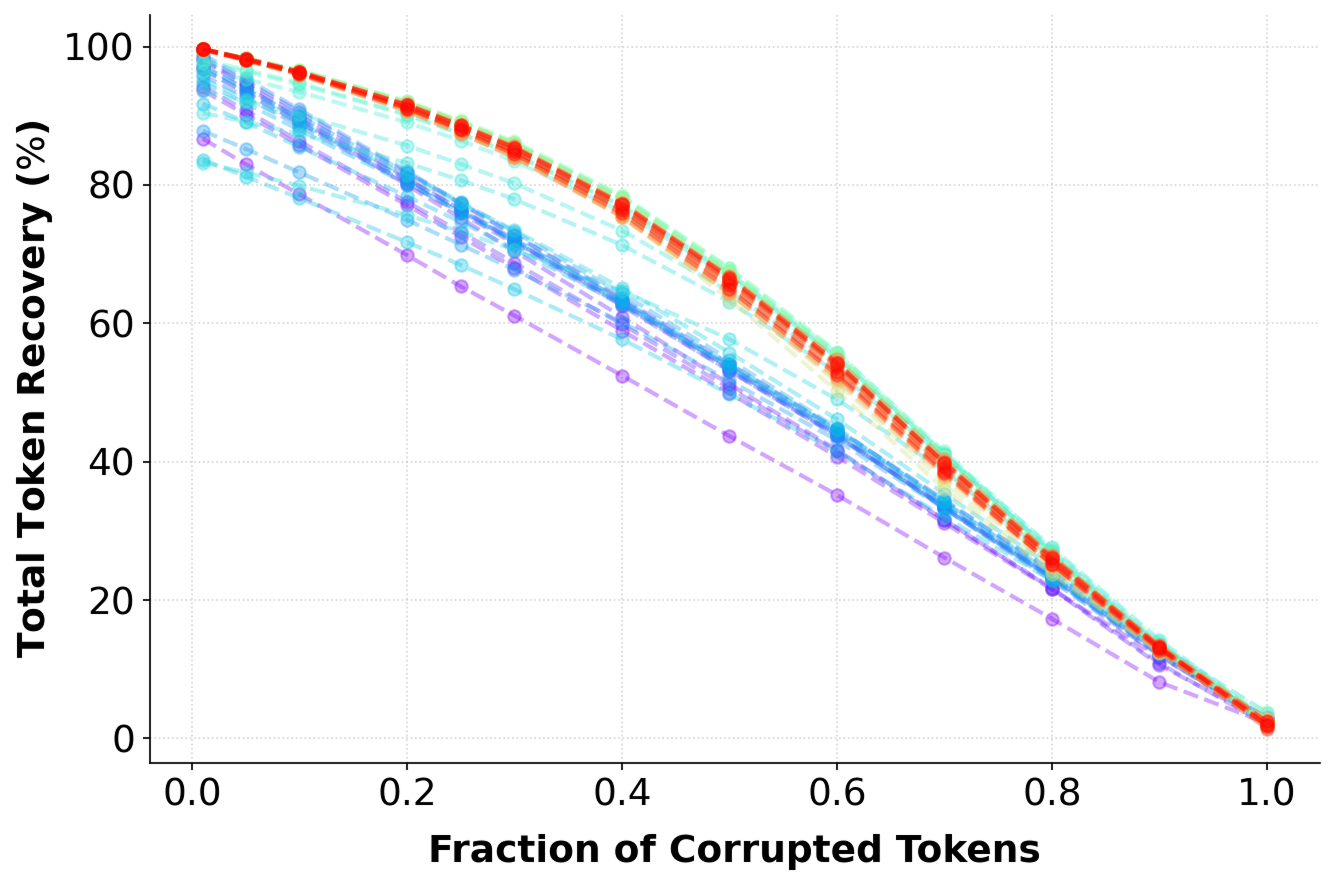}
             \caption{\fontsize{12}{16}\selectfont Test}
        \end{subfigure}
        \vspace{-3mm}
        %\captionof*{figure}{Medium ($\sim$ 384\textbf{M})} 
        \par{Medium ($\sim$ 384\textbf{M})}
    \end{minipage}
    \vspace{-1mm}
    \caption{\textbf{Shrinkage and expansion of training and test samples' basins of attraction during the memorization-to-generalization transition for various UDDM sizes}. Total token recovery rates (\%), including non-corrupt and corrupted, are shown across varying levels of corruption for (\textbf{a}) training sequences and (\textbf{b}) unseen test samples. The color gradient represents the fraction of the training dataset used, ranging from \textcolor[rgb]{0.4, 0.0, 0.6}{small} to \textcolor{red}{large}. As the training dataset grows, the recovery rate for training examples diminishes, reflecting a shrinkage in the basins of attraction around explicitly memorized points. Simultaneously, the recovery rate for unseen test samples improves, demonstrating an expansion of the basins of attraction around novel examples within the broader data distribution. The convergence of these recovery rates at large dataset sizes (red curves) signifies that unseen test samples have effectively become stable attractors, marking the shift from pure memorization to generalization.}
    \label{fig:basin_shrinkage_extra}
\end{figure}

\begin{figure}[H]
    \centering
    \begin{minipage}{1\textwidth}
        \begin{subfigure}[b]{1.0\textwidth}
            \centering
            \includegraphics[width=\linewidth]{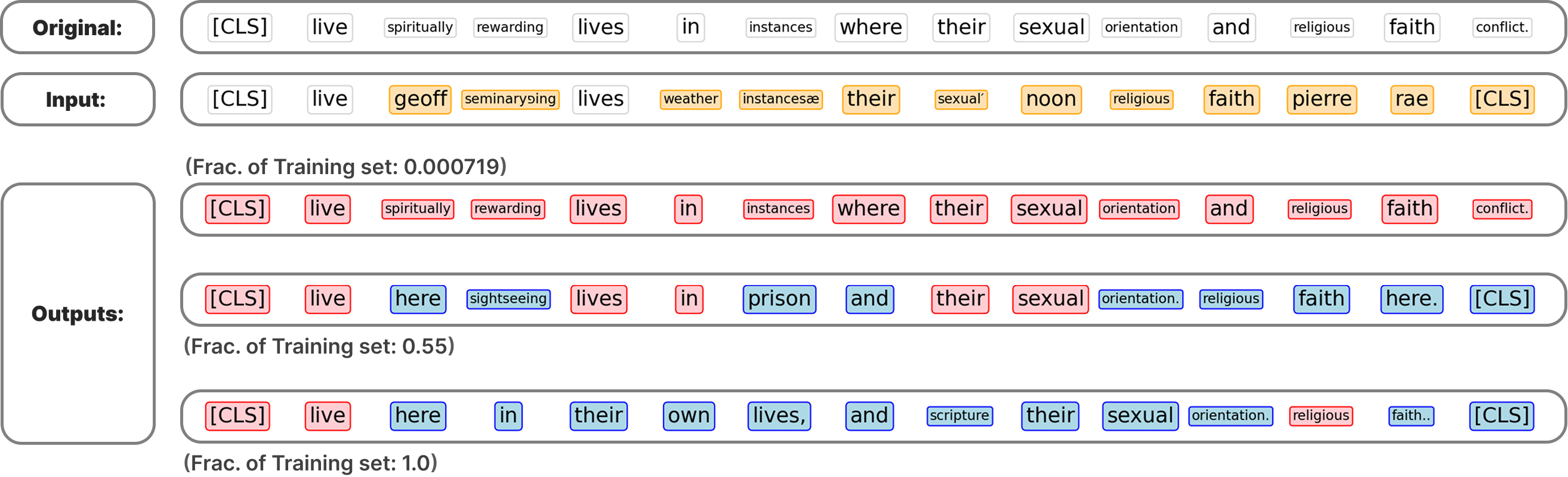}
            \caption{\fontsize{11}{12}\selectfont Tiny ($\sim$ 24\textbf{M})}
        \end{subfigure}
    \end{minipage}
    \vskip 0.35cm
    \begin{minipage}{1\textwidth}
        \begin{subfigure}[b]{1.0\textwidth}
            \centering
            \includegraphics[width=\linewidth]{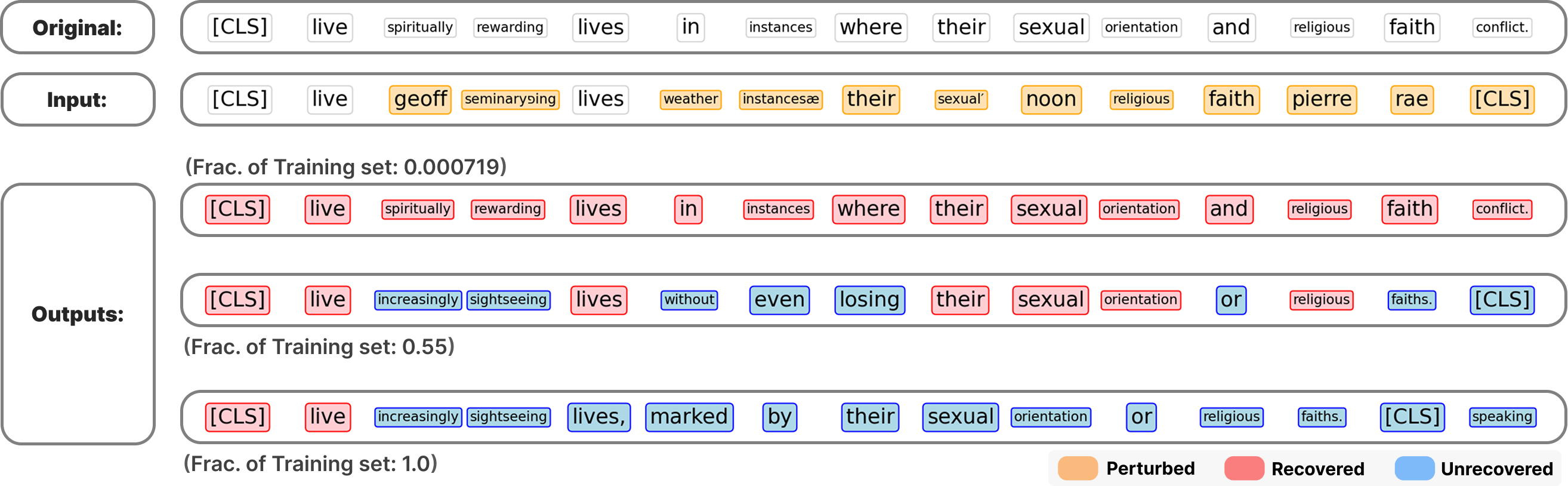}
            \caption{\fontsize{11}{12}\selectfont Small ($\sim$ 135\textbf{M})}
        \end{subfigure}
    \end{minipage}
    \vskip 0.35cm
    \begin{minipage}{1\textwidth}
        \begin{subfigure}[b]{1.0\textwidth}
            \centering
            \includegraphics[width=\linewidth]{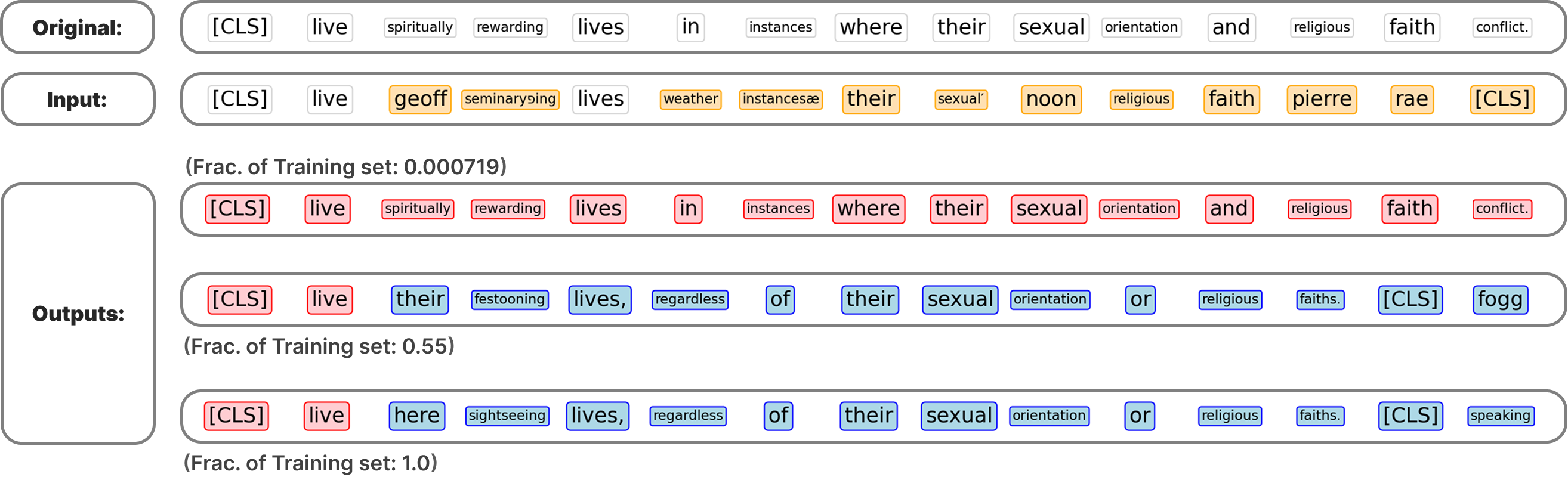}
            \caption{\fontsize{11}{12}\selectfont Medium ($\sim$ 384\textbf{M})}
        \end{subfigure}
    \end{minipage}
    %\vskip -0.25cm
    \caption{An illustration of the model's ability to recover tokens from perturbed sequences on \textbf{training examples} at three different fractions of the training dataset and sizes of the UDDMs. Perturbation is computed at $t = 0.5$ and the typical stochastic reverse process is performed afterwards. As the training dataset size increases, the model's ability to recover perturbed tokens becomes worse and in contrast, its generative ability improves. Meanwhile, there is no distinction among the various UDDMs' sizes, where the overall trend of memorization and generalization in relation to the training dataset size persists. }
    \label{fig:train_recovery_examples}
\end{figure}

\begin{figure}[htbp]
    \centering
    \begin{minipage}{1\textwidth}
        \begin{subfigure}[b]{1.0\textwidth}
            \centering
            \includegraphics[width=\linewidth]{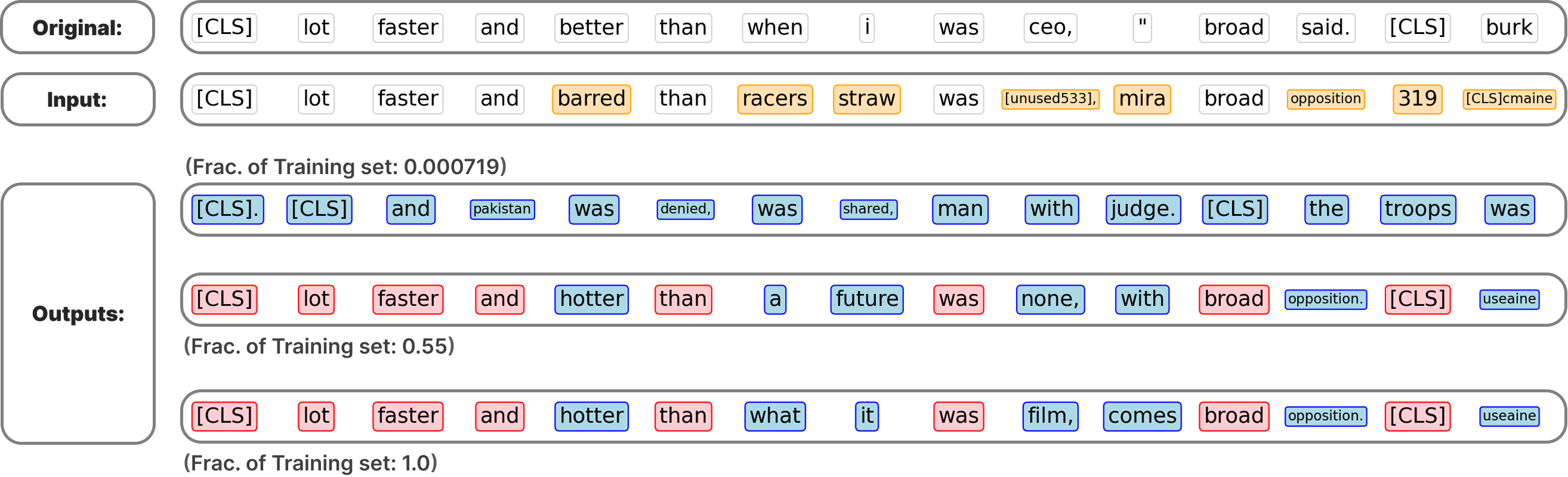}
            \caption{\fontsize{11}{12}\selectfont Tiny ($\sim$ 24\textbf{M})}
        \end{subfigure}
    \end{minipage}
    \vskip 0.35cm
    \begin{minipage}{1\textwidth}
        \begin{subfigure}[b]{1.0\textwidth}
            \centering
            \includegraphics[width=\linewidth]{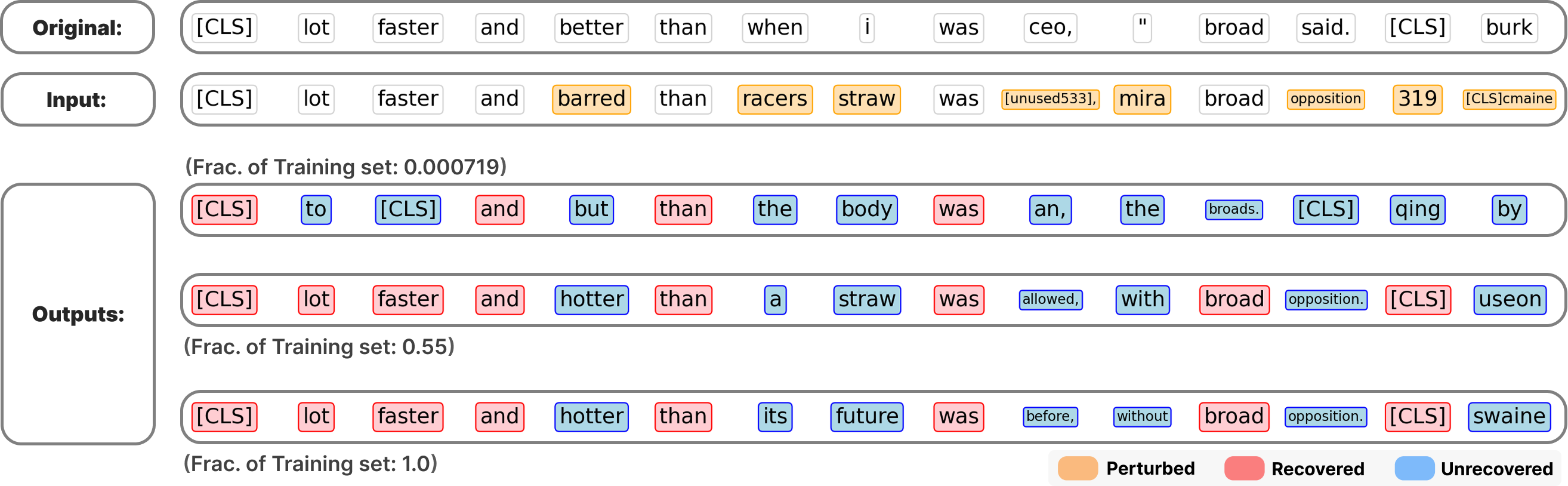}
            \caption{\fontsize{11}{12}\selectfont Small ($\sim$ 135\textbf{M})}
        \end{subfigure}
    \end{minipage}
    \vskip 0.35cm
    \begin{minipage}{1\textwidth}
        \begin{subfigure}[b]{1.0\textwidth}
            \centering
            \includegraphics[width=\linewidth]{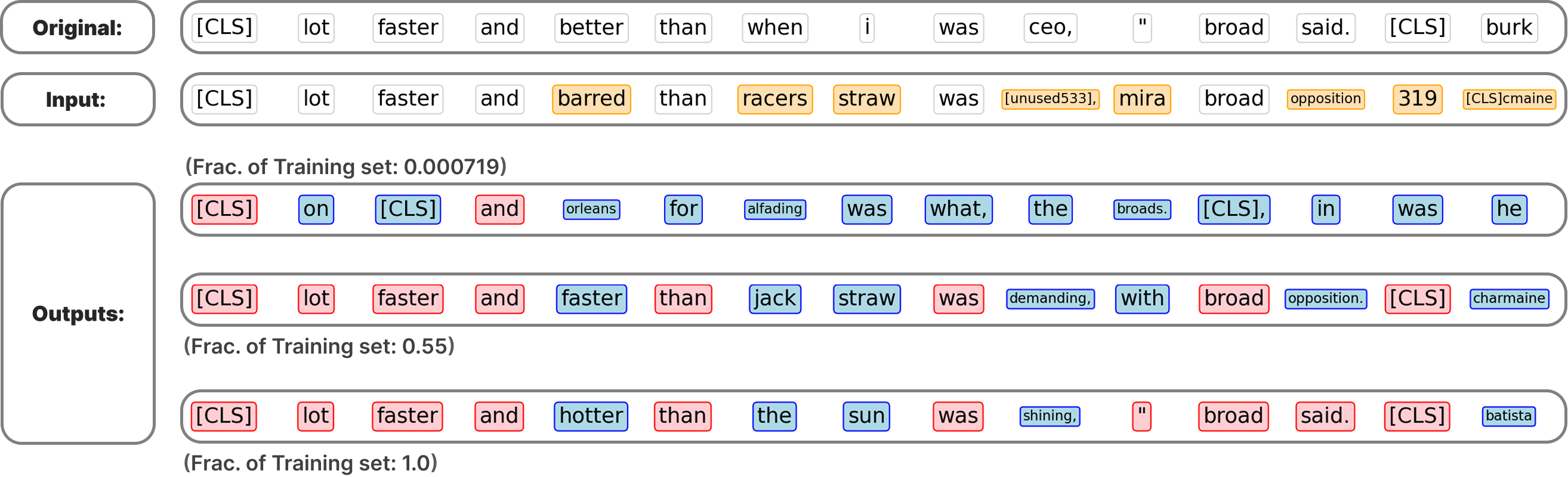}
            \caption{\fontsize{11}{12}\selectfont Medium ($\sim$ 384\textbf{M})}
        \end{subfigure}
    \end{minipage}
    %\vskip -0.15cm
    \caption{An illustration of the model's ability to recover tokens from perturbed sequences on \textbf{test examples} at different fractions of the training dataset and sizes of UDDMs. Perturbation is computed at $t = 0.5$ and the stochastic reverse process is performed afterwards. In the beginning, the model is unable to recognize unperturbed unseen test tokens, where it often change them to another tokens. However, as the training dataset size increases, the UDDM is more likely to maintain the unperturbed test tokens rather than `flipping' them. 
    %
    %Moreover, as one increases the model's size, the stability of unperturbed and unseen test tokens seems to be increased (during memorization or when the training dataset size is small), reflecting very well with \cref{fig:basin_shrinkage_extra}.
    }
    \label{fig:test_recovery_examples}
\end{figure}

\smallskip
\noindent \textbf{Textual Examples}. In Figs.~(\ref{fig:basin_shrinkage}) and (\ref{fig:basin_shrinkage_extra}), we are interested in studying the stability of unseen test tokens across different perturbation levels (although we are much more interested in the case where there is little noise for these experiments). For the textual examples illustrated in Figs.~(\cfig{fig:basin_shrinkage}{A}) and (\cfig{fig:unrecovered_recovered}{A}), we used the models that have been trained with the fractions of the training dataset: $0.000719$ and $1.0$. Although these two figures utilize different sampling dynamics (i.e., greedy versues stochastic), in general we do not spot much differences between the greedy and stochastic processes for our token recovery experiments. Please see Figs.~(\cfig{fig:unrecovered_recovered}{A}), (\ref{fig:train_recovery_examples}), and (\ref{fig:test_recovery_examples}) for the textual examples that collected using the typical stochastic process.

%To avoid stochastic noise during the sampling process, we used the greedy process by replacing $\text{Cat}(\cdot)$ with the $\argmax$ operation in \cref{eqn:discrete-reverse-process} to obtain something similar to the deterministic process (\ref{eq:deterministic_update}) in the simple setting detailed in \cref{sec:am_pseudo}. 

\smallskip
\noindent \textbf{Perplexity}. Similarly to our results on conditional entropy, we utilized the same sets of $10^{5}$ generated sequences belonging to each trained UDDM, respectively. The code to compute perplexity is also from \citet{sahoo2025duality}. Keep in mind, the reported conditional entropy results are computed from our UDDMs, unlike perplexity, which requires the reference or critic model of the pre-trained GPT2-Large \cite{radford2019language} as mentioned previously.

\smallskip
\noindent \textbf{Hardware}. The training of the UDDMs are done using NVIDIA Tesla V100 GPUs. Each GPU has 32GB of memory and is linked with Power9 processors, clocking at 3.15 GHz maximum. For each model, we used 4 GPUs and an effective total (or global) batch size of $512$ samples. For each GPU, the local batch size is set as $64$, requiring 2 gradient accumulation steps.

\section{Usage Of LLM Disclaimer}
LLMs were utilized in part of the writing of the manuscript, specifically for fine-tuning sentences and grammar correction.

\onecolumn
\section{Uniform-state Discrete Diffusion and Duality with Gaussian}
\label{sec:duo}

\textbf{Further Details on UDDMs}. As mentioned in \cref{sec:prelim}, the true reverse posterior of a previous timestep $s < t$ corresponding to the forward process (\ref{eqn:discrete-forward-process}) is
\begin{equation}
\begin{split}
     \rvz_s \sim q_{s | t}(\rvz_s | \rvz_t, \rvx) &= \text{Cat} \Big (\rvz_s; g(\rvz_t, \rvx) \Big ) \\
     &= \text{Cat} \Bigg (\rvz_s ; \frac{K \alpha_t \rvz_t \odot \rvx + (\alpha_{t|s} -\alpha_t)\rvz_t}{K \alpha_t \langle \rvz_t, \rvx \rangle + (1 - \alpha_t)} + \frac{(\alpha_s - \alpha_t)\rvx + (1 - \alpha_{t | s}) \frac{\mathbf{1}}{K}}{K \alpha_t \langle \rvz_t, \rvx \rangle + (1 - \alpha_t)} \Bigg ), 
\end{split}
\label{eqn:discrete-reverse-process-full}
\end{equation}
where $\langle \cdot, \cdot \rangle$ denotes the dot product, $\odot$ denotes Hadamard product, the relative diffusion parameter is $\alpha_{t|s} =\frac{\alpha_t}{\alpha_s}$, and we have a uniform prior over $\mathcal{V} (\boldpi = \mathbf{1} / {K})$ \citep{sahoo2024simple, sahoo2025duality}.

Since the true clean token $\rvx$ is unknown during the generative process (\ref{eqn:discrete-reverse-process-full}), the approximate reverse posterior is defined as $p^\theta_{s|t}(\rvz_s | \rvz_t) = q_{s | t} (\rvz_s | \rvz_t, \rvx = \rvx_\theta(\rvz_t, t))$. The neural network $\rvx_\theta(\rvz_t, t) \approx \rvx$ is trained to predict the clean token $\rvx$ at any time $t$ according to \cref{eqn:discrete-reverse-process}, where $p_\theta (\rvx | \rvz_t) = \rvx_\theta (\rvz_t, t)$. The network's parameters $\theta$ are optimized via the Negative Evidence Lower Bound (NELBO) objective \citep{austin2021structured, sahoo2024simple}: 
\begin{equation}
\mathcal{L}_\text{\tiny NELBO}=
    \mathbb{E}_q \Big [
    \underset{\mathcal{L}_\text{reconstruction}}{\underbrace{- \log p_\theta (\rvx | \rvz_0)}} + \underset{\mathcal{L}_\text{diffusion}}{\underbrace{\sum_{s < t} D_\text{KL} \big [ q(\rvz_s | \rvz_t , \rvx) \, || \, p_\theta (\rvz_s | \rvz_t ) \big ]}}
    \Big ] + \underset{\mathcal{L}_\text{prior}}{\underbrace{D_\text{KL} \big [ q(\rvz_1 | \rvx) \, || \,  p_\theta (\rvz_1) \big ]}}.
    \label{eqn:nelbo-full}
\end{equation}
As discussed in \cref{sec:am_pseudo}, the term $\mathcal{L}_\text{reconstruction}$ is more related to our simple setting in that section due to the fact that it only deals with a single finite temperature value (or it does not rely on temperature annealing). However, we suspect that the other terms in NELBO (\ref{eqn:nelbo-full}) further help the maximization of classification margin (or basin) of the data points.

\medskip \noindent 
\textbf{Duality of Uniform and Gaussian}. In continuous diffusion modeling, we typically rely on the diffusion mapping of a data distribution $q_\text{data}$ to a simple prior distribution that is often the standard Gaussian distribution $\mathcal{N}(0, \mathbf{I}_K)$. The marginal distribution of the noisy latent variable $\rvw_t \sim \tilde{q}_t(\cdot | \rvx)$ at time $t$ is defined as:
\begin{equation}
    \rvw_t \sim \tilde{q}_t(\rvw_t | \rvx; \tilde{\alpha}_t) = \mathcal{N} (\rvw_t ;  \tilde{\alpha}_t \rvx , (1 -\tilde{\alpha}_t^2) \mathbf{I}_K),
\end{equation}
where $\tilde{\alpha}_t \in [0, 1]$ is the diffusion parameter that is a monotonically decreasing function in $t$. The boundary conditions are $\tilde{q}_{t = 0} \approx q_\text{data}$ and $\tilde{q}_{t = 1} = \mathcal{N}(0, \mathbf{I}_K)$.

However, as shown in \citet{sahoo2025duality}, there exists a connection between the Gaussian and Uniform diffusion processes for the discrete setting. Specifically, we can utilize the operator, $\argmax : \mathbb{R}^K \rightarrow \mathcal{V}$, to map a continuous vector $\rvw \in \mathbb{R}^K$ to the one-hot vector corresponding to $\argmax (\rvw) = \underset{\rvz \in \mathcal{V}}{\argmax} \,\, \rvz^\top \rvw$. Then, we can define the discrete marginals to be 
\begin{equation}
\rvz_t = \argmax ( \rvw_t ),
\end{equation}
and the conditional probability mass function $p_t(\rvz_t | \rvx)$ marginalized over $\rvw_t \sim \tilde{q}_t (\rvw_t | \rvx; \tilde{\alpha}_t )$ such that 
\begin{equation}
    \rvz_t \sim P_t \big(\rvz_t | \rvx; \mathcal{T} (\tilde{\alpha}_t) \big ) = \Cat \bigg ( \rvz_t ; \mathcal{T}(\tilde{\alpha}_t) \rvx + (1 - \mathcal{T}(\tilde{\alpha}_t) ) \frac{\mathbf{1}}{K} \bigg ) ,
\label{eqn:discrete-marginal-dual}
\end{equation}
where $\mathcal{T} : [0, 1] \rightarrow [0, 1]$ is the Gaussian Diffusion Transformation operator. This operator is defined as the following Gaussian integral:
\begin{equation}
    \alpha_t = \mathcal{T}(\tilde{\alpha}_t) = \frac{K}{K - 1} \bigg [ \int^\infty_{-\infty} \phi \bigg ( z - \frac{\tilde{\alpha}_t}{\sqrt{1 - \tilde{\alpha}^2_t}} \bigg ) \Phi^{K - 1} (z) \dd z - \frac{1}{K} \bigg ]
\end{equation}
where $\phi(z) = \frac{\exp (-z^2)}{\sqrt{2 \pi}}$ is the standard Normal distribution and $\Phi(z) = \int^z_{-\infty} \phi(t) \dd t$ is the respective cumulative distribution.

%The time evolution of the marginals $\rvz_t$ remains similar to \cref{eqn:uniform-evol-marg}, but it involves the operator $\mathcal{T}(\tilde{\alpha}_t)$ in the linear ODE: 
%\begin{equation}
%    \frac{\dd}{\dd t} P_t = - \frac{-\mathcal{T}'(\tilde{\alpha}_t)}{K \mathcal{T}(\tilde{\alpha}_t}  \big [ \mathbf{1} \mathbf{1}^\top - K \mathbf{I}_K \big ] P_t
%\label{eqn:uniform-evol-marg-dual}
%\end{equation}
%where $\mathcal{T}'(\tilde{\alpha}_t)$ denotes the time derivative of $\mathcal{T}(\tilde{\alpha}_t)$. 

%A fundamental connection is established between Uniform-state discrete and Gaussian diffusion processes after inspecting Eqs.~(\ref{eqn:discrete-marginal-dual}) and (\ref{eqn:uniform-evol-marg-dual}). Specifically, the $\argmax$ operator transforms the Gaussian diffusion process into Uniform-state diffusion process. Moreover, \cite{sahoo2025duality} expresses this formal connection as
Overall, there exists a fundamental connection between Uniform-state discrete and Gaussian diffusion processes, shown in \citet{sahoo2025duality}. Specifically, this formal connection is expressed as 
\begin{equation}
    \rvz_t \sim q_t(\rvz_t | \rvx; \mathcal{T}(\tilde{\alpha}_t)) = [\argmax]_{\boldsymbol{*}} \,\tilde{q}_t (\rvw_t | \rvx ; \tilde{\alpha}_t)
    \label{eqn:forward-duo}
\end{equation}
where $\boldsymbol{*}$ denotes the push-forward of the $K$-dimensional Gaussian density $\tilde{q}_t$ under $\argmax$ which yields a categorical distribution of $K$ categories. For full details, please refer to \cite{sahoo2025duality}.

\section{Conditional Entropy and Curvature}
\label{sec:entropy_curvature}
In the more common setting, where AM have an energy landscape, the sharpness of the basin dictates retrieval dynamics \citep{DenseAM2016, KrotovFrontiers, krotov2025modern}. Low entropy implies more deterministic attractors (\textit{memorization}), while high entropy signals a flatter landscape with distributed probability mass \citep{biroli2024dynamical}. This flatness facilitates \textit{generalization}, where basins widen and merge to capture the underlying data manifold, enabling the synthesis of novel patterns \citep{pham2025memorization}.

In this section, inspired by \citet{biroli2024dynamical} and \citet{d2025pseudo}, we relate entropy and the curvature (or sharpness) of the energy in the continuous setting, using local approximation, to show the connection between these two ideas. Here, assume that the clean data $\rvx \in \mathbb{R}^d$ and its perturbed version $\rvz_t \in \mathbb{R}^d$ at time $t$.

\noindent \textbf{Proof Sketch}. Consider the conditional distribution $p(\rvx | \rvz_t)$ defined by an energy function $E(\rvx ; \rvz_t)$:
\begin{equation}
    p(\rvx | \rvz_t) = \frac{1}{Z(\rvz_t)} e^{-E(\rvx ; \; \rvz_t)},
\end{equation}
where $Z(\rvz_t) = \int e^{-E(\rvy, \rvz_t)} \dd \rvy$ is the partition function. We assume the distribution is peaked around a mode $\rvx^*$, representing the most likely clean data point given the noisy observation $\rvz_t$.

To analyze the local geometry, we perform a second-order Taylor expansion of the energy $E(\rvx; \rvz_t)$ with respect to $\rvx$, centered around the mode $\rvx^*(\rvz_t)$:
\begin{equation}
    E(\rvx; \rvz_t) \approx E(\rvx^*; \rvz_t) + (\rvx - \rvx^*)^\top \nabla_{\rvx} E(\rvx^*; \rvz_t) + \frac{1}{2} (\rvx - \rvx^*)^\top \mathbf{H}(\rvz_t) (\rvx - \rvx^*).
    \label{eqn:taylor-expan-cond-energy-min}
\end{equation}
Since $\rvx^*$ is a local minimum of the energy surface defined by $\rvz_t$, the gradient $\nabla_{\rvx} E(\rvx^*; \rvz_t)$ vanishes. The matrix $\mathbf{H}(\rvz_t) = \nabla_{\rvx}^2 E(\rvx^*; \rvz_t)$ is the Hessian of the energy, representing the local curvature or \textit{sharpness} of the energy basin conditioned on $\rvz_t$.

Using \cref{eqn:taylor-expan-cond-energy-min} and assuming that we are at the minimum where $\nabla_\rvx E(\rvx^*; \rvz) = \mathbf{0}$, we can perform Laplace approximation for the partition function $Z(\rvz_t)$: 
\begin{equation}
    \begin{split}
    Z(\rvz_t) \approx & \int e^{- \big( E(\rvx^*; \rvz_t) + \frac{1}{2} (\rvx - \rvx^*)^\top \mathbf{H}(\rvz_t) (\rvx - \rvx^*) \big )} \dd \rvx\\
    &= e^{-E(\rvx^*; \rvz_t )}  \int e^{-\frac{1}{2} (\rvx - \rvx^*)^\top \mathbf{H}(\rvz_t) (\rvx - \rvx^*) } \dd \rvx \\ 
    &= (2 \pi)^\frac{d}{2} \, e^{-E(\rvx^*; \rvz_t )} \, \mathrm{det} \big (\mathbf{H}(\rvz_t) \big )^{-\frac{1}{2}}
    \end{split}
\end{equation}
The conditional entropy $\mathcal{H} (\rvx | \rvz_t)$ can now be defined. Using the relationship $\log p = -E - \log Z$ and substitute it into $\mathcal{H} (\rvx | \rvz_t)$, we have 
\begin{equation}
\begin{split}
\mathcal{H} (\rvx | \rvz_t) & = \underset{\rvx \sim p(\rvx | \rvz_t)} {\mathbb{E}} \bigg [ E (\rvx ; \rvz_t) + \log Z(\rvz_t) \bigg ] \\ 
& = \underset{\rvx \sim p(\rvx | \rvz_t)} {\mathbb{E}} \bigg [ E(\rvx; \rvz_t) + \frac{d}{2} \log (2 \pi) - E(\rvx^* ; \rvz_t ) - \frac{1}{2} \log \big [ \mathrm{det}\big(\mathbf{H} (\rvz_t) \big) \big ]  \bigg ]
\end{split}
\label{eqn:entropy-pf1}
\end{equation}
If we substitute \cref{eqn:taylor-expan-cond-energy-min} into the term $E(\rvx; \rvz_t)$ in \cref{eqn:entropy-pf1}, we then have 
\begin{equation}
\begin{split}
 \mathcal{H} (\rvx | \rvz_t) & \approx \underset{\rvx \sim p(\rvx | \rvz_t)} {\mathbb{E}} \bigg [ \cancel{E(\rvx^* ; \rvz_t)} +   \frac{d}{2} \log (2 \pi) \cancel{- E(\rvx^* ; \rvz_t )} - \frac{1}{2} \log \big [\mathrm{det}\big(\mathbf{H} (\rvz_t) \big) \big ] + C \bigg ] \\ 
 & \approx \underset{\rvx \sim p(\rvx | \rvz_t)} {\mathbb{E}} \bigg [ - \frac{1}{2} \log \big [\mathrm{det}\big(\mathbf{H} (\rvz_t) \big) \big ] +C \bigg ] %= - \frac{1}{2} \log \big [\mathrm{det}\big(\mathbf{H} (\rvz_t) \big) \big ] +C
\end{split} 
\end{equation}
where $C$ is a constant involving the omitted terms from our substitution of \cref{eqn:taylor-expan-cond-energy-min}. \\

\noindent \textbf{Discussion}. Overall, this derivation highlights that the conditional entropy is inversely proportional to the log-determinant of the Hessian at the mode, and aligns well to the findings of \citet{biroli2024dynamical} where a collapse in entropy corresponds to the system getting trapped in small-disjoint regions of the configuration space. However, in this work, we are exploring UDDMs, which are not continuous DMs. Thus, we lack the formulations that attempt to link up their conditional entropy with the sharpness in the discrete setting of language or text modeling. But we suspect that the connection between Uniform and Gaussian distributions in the discrete setting, laid out by \citet{sahoo2025duality}, provides some clues to further this link between conditional entropy and energy in a future work.  

\begin{figure}[t]
    \centering
    \includegraphics[width=0.49\linewidth]{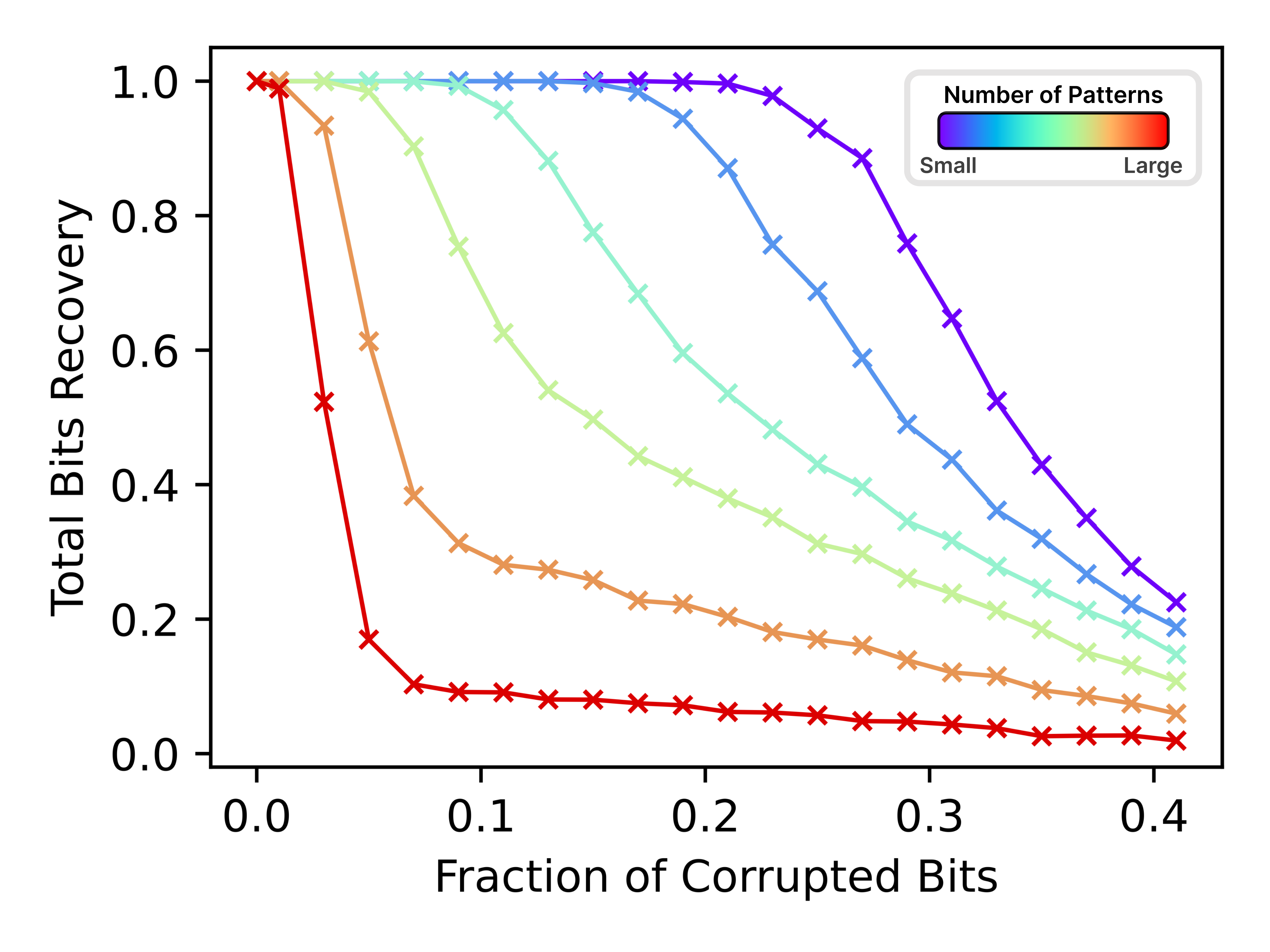}
    \includegraphics[width=0.49\linewidth]{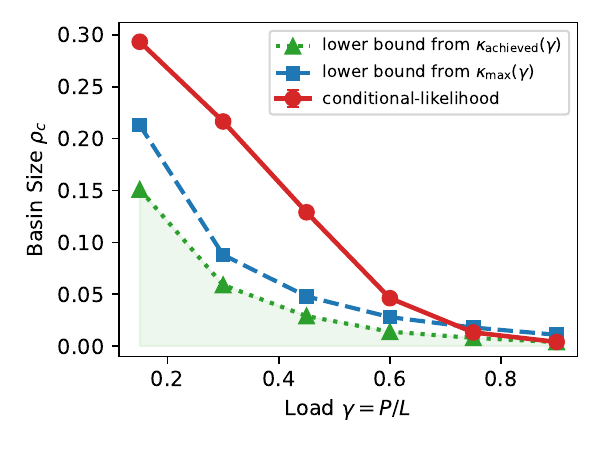}
    \caption{
    \textbf{The implicit maximization of the margin guarantees large basins of attraction.} Both panels refer to an associative memory model trained with conditional likelihood on random bit sequences. \textit{Left Panel:} Average total bit recovery rates for training sequences across varying corruption levels. Line colors indicate the sizes of the training datasets used. This panel captures the same qualitative memorization phenomenon of \cref{fig:basin_shrinkage}. \textit{Right Panel:} We plot the basin size $\rho_c(\gamma)$ of the trained model (\textcolor{red}{red} curve) compared to the margin-derived lower bound, evaluated both at the idealized $\kappa_{\max}(\gamma)$ and at the empirically achieved $\kappa_{\text{achieved}}(\gamma)$ (\textcolor{blue}{blue} and \textcolor{green}{green} curves respectively). The measured basin size remains above the achieved-margin bound at every load, confirming margin maximization is sufficient to produce basins of the predicted size. The red points correspond to the critical noise level for which the token recovery rate drops below 0.99. Each point is the average of five samples, the error bars are invisible.}
    \label{fig:basin_verification}
\end{figure}

\section{Numerical Validation of Section (\ref{sec:am_pseudo})}
\label{sec:numerical_validation_of_PL}

In this section we propose a numerical verification of the binary or linear-logit analysis of Sec.~\ref{sec:am_pseudo}, specifically that optimizing the conditional likelihood in \cref{eqn:pseudo-likelihood} implicitly leads to a maximum-margin solution, which in turn leads to basins of attraction around training data points. Please note that these results are all already established \cite{Gardner1988OptimalSP, Soudry2018, d2025pseudo}, and that we reproduce them here for convenience.\\

\noindent \textbf{Setup.}
We generate $P = \gamma L$ random i.i.d.\ binary sequences (or patterns), where each sequence has a length of $L = 250$, and train the recurrent coupling matrix $\mathbf{W}$ by minimizing the pseudo-likelihood loss of \cref{eqn:pseudo-likelihood} for each row independently (unregularized logistic regression), which is known to converge in direction to the maximum-margin solution \citep{Soudry2018}. Here, $\gamma = \frac{P}{L}$ is the load or the ratio between the number of patterns $P$ and their dimension $L$. For each load $\gamma$, we then measure the network's empirical basin of attraction: starting from a stored pattern, we flip a fraction $\rho$ of its bits, run the recurrent dynamics to a fixed point, and record the mean overlap between the recovered state and the original pattern, averaged over multiple patterns and corruption realizations. Basin size $\rho_c(\gamma)$ is defined as the largest corruption level for which this overlap remains above $0.99$, following the same criterion used in \citet{d2025pseudo}. The results are shown in \cref{fig:basin_verification}. \\

\noindent \textbf{Theoretical comparison.}
We compare $\rho_c(\gamma)$ against the noise tolerance implied by the Gardner capacity \citep{Gardner1988OptimalSP}:
\begin{equation}
\frac{1}{\gamma_c(\kappa)} = (1+\kappa^2)\,\Phi(\kappa) + \kappa\,\phi(\kappa),
\label{eq:gardner_capacity}
\end{equation}
with $\Phi$ the cumulative distribution function and $\phi$ the probability density function of the standard normal distribution, inverted numerically to obtain $\kappa_{\max}(\gamma)$.
Here $\kappa$ is normalized so that a neuron's pre-activation has unit variance under random patterns, i.e.\ $\kappa$ is a signal-to-noise ratio (a $z$-score). Corrupting a fraction $\rho$ of the $N$ inputs perturbs the pre-activation by a sum of $\rho N$ independent random contributions, which by the central limit theorem has standard deviation $\sim 2\sqrt{\rho}$. A stored pattern with margin $\kappa$ therefore remains a fixed point as long as the injected noise does not exceed the margin, i.e.\ $\kappa > 2\sqrt{\rho}$. This gives a closed-form, margin-derived lower bound on basin size, which we plot alongside the measured $\rho_c(\gamma)$.\\

\noindent \textbf{Slow convergence to $\kappa_{\max}$.}
In practice, the margin actually achieved by finite-time training, $\kappa_{\text{achieved}}(\gamma)$, undershoots the asymptotic optimum $\kappa_{\max}(\gamma)$, since convergence to the max-margin direction under logistic loss is logarithmic in training time \citep{Soudry2018}, a phenomenon also reported in \citet{d2025pseudo}. We therefore report both curves: the lower bound computed from the idealized $\kappa_{\max}(\gamma)$, and the one computed from the empirically achieved $\kappa_{\text{achieved}}(\gamma)$, which is the fair comparison for a network trained for a finite number of steps.

\noindent \textbf{Result.}
\cref{fig:basin_verification} shows that the measured basin size lies consistently above the bound derived from $\kappa_{\text{achieved}}(\gamma)$ at every load tested, confirming that the margin does provide a quantitative guarantee on basin size. This directly supports the mechanism claimed in Sec.~(\ref{sec:am_pseudo}): maximizing the margin is sufficient to produce basins of attraction of predictable size.

\section{Additional Results}
\label{sec:add-results}
\input{EMNLP/figures/train_synth_cond}

%\clearpage

\subsection{Visualizations of Text Recovery Examples}
\label{sec:text_recovery_examples}

\subsection{Conditional Entropy Histograms}
\label{sec:additional-hists}

\begin{figure}[h]
    \centering
    \includegraphics[width=1\linewidth]{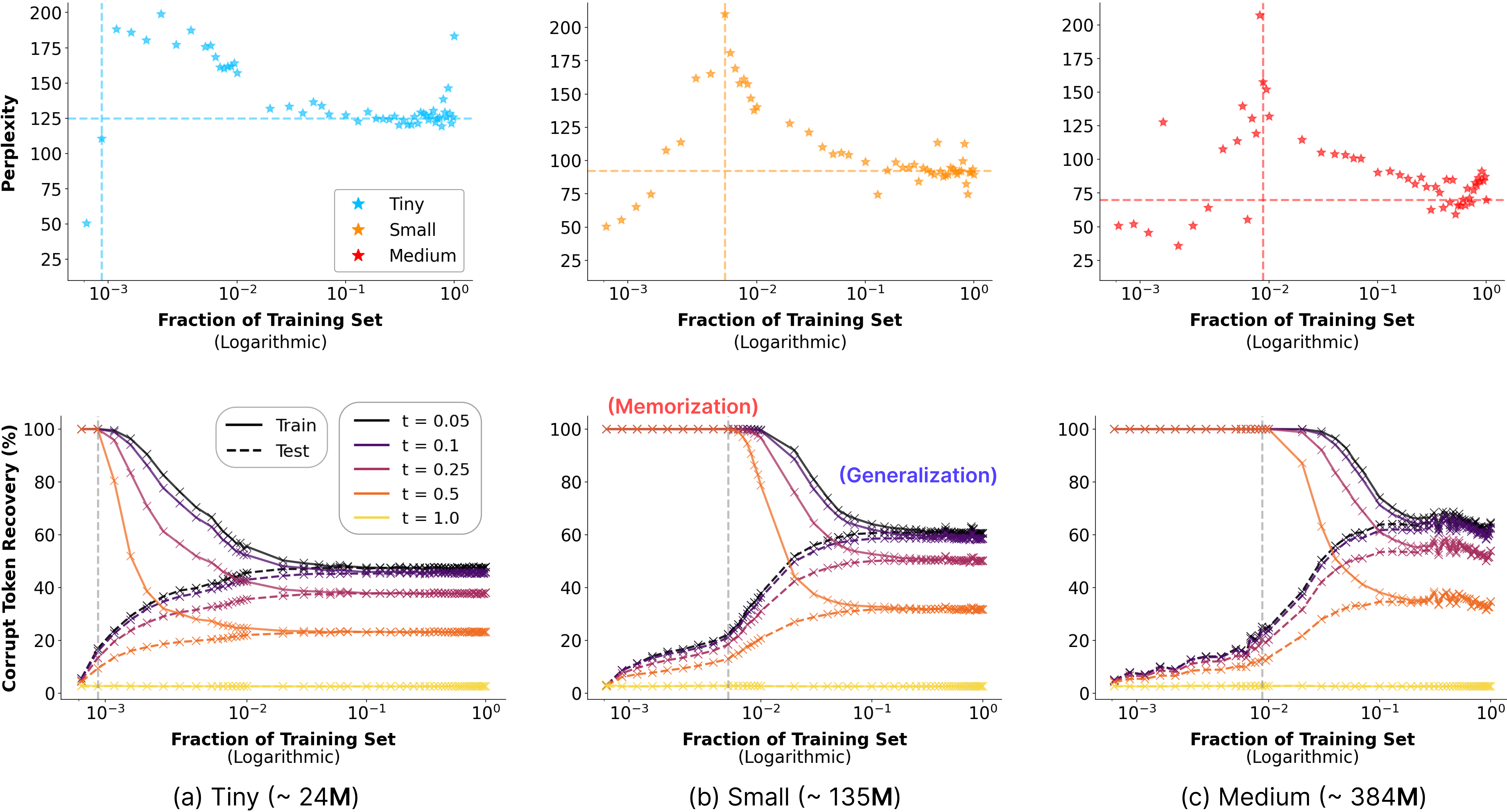}
    \caption{\textbf{Perplexity trends during the memorization-to-generalization transition together with corrupted token recovery}. This figure demonstrates the alignment of phase transitions across perplexity (top) and the average corrupt token recovery rate (bottom) for three model sizes. The \textit{vertical} dashed lines roughly represent the critical transitional points where the perplexity (calculated using GPT-2 Large \cite{radford2019language}) aligns with the initial decrease of corrupt token recovery rate for training and the increase of test tokens' recovery rate. This alignment of the two metrics showcases the usefulness of the token recovery rate in determining whether generalization has been reached in UDDMs.
    }
    \label{fig:ppl_combined}
\end{figure}

%% file: EMNLP/figures/train_synth_cond.tex
\begin{figure}[htbp]
    \centering
    \includegraphics[width=1.0\linewidth]{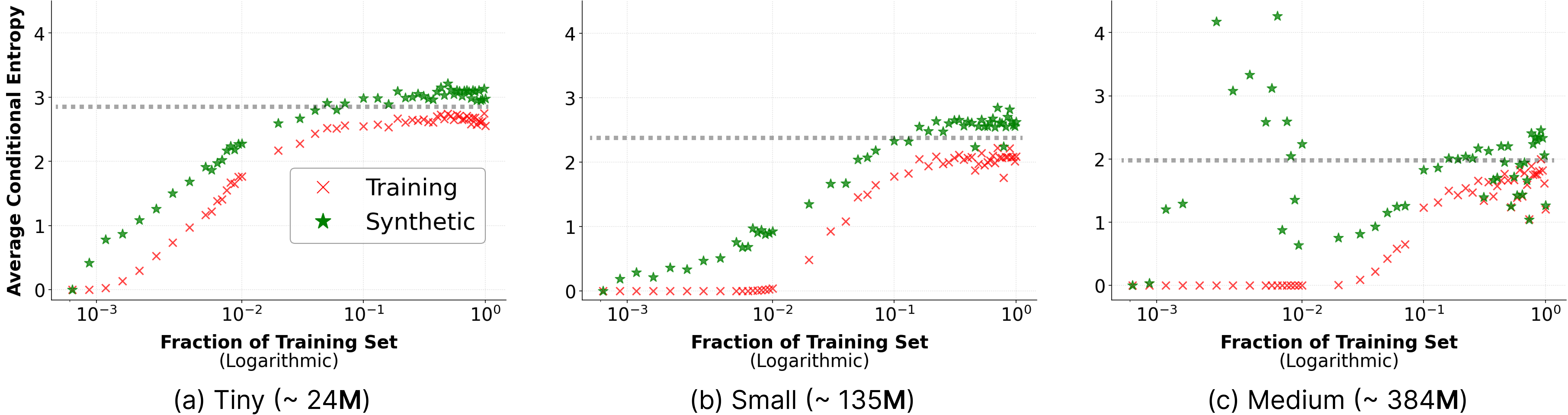}
    \caption{\textbf{Average conditional entropy for training versus synthetic sequences}. As the training dataset size increases, the average (over sequences and tokens) conditional entropy for both \textcolor{red}{training} and \textcolor[rgb]{0, 0.6, 0.2}{synthetic} (or generated) sequences naturally rises. Here, we plot the dashed horizontal line in each plot to show the location of the separation between the average conditional entropies of training and synthetic sequences. Initially, there exists an ``entropy gap'' which separates these two sample types given small fractions of the training set, depicting the UDDM's uncertainty during generation. However, scaling up the model's size narrows this entropy gap, demonstrating that higher parameter counts increase the model's confidence in its own generated text during the generalization regime.}
    \label{fig:entropy_over_time}
\end{figure}